\documentclass{article}

\usepackage{PRIMEarxiv}
\usepackage{pythonhighlight}

\usepackage[utf8]{inputenc} 
\usepackage[T1]{fontenc}    
\usepackage[hyperfootnotes=false]{hyperref}
\usepackage{url}            
\usepackage{booktabs}       
\usepackage{amsfonts}       
\usepackage{nicefrac}       
\usepackage{microtype}      
\usepackage{lipsum}
\usepackage{fancyhdr}       
\usepackage{graphicx}       
\graphicspath{{media/}}     
\usepackage{comment}
\usepackage{amsmath,amssymb}
\usepackage{appendix}
\usepackage{multirow}
\usepackage{booktabs}
\usepackage{tablefootnote} 
\usepackage{subcaption}
\usepackage{authblk}
\usepackage{hyperref}

\newcommand\blfootnote[1]{%
  \begingroup
  \renewcommand\thefootnote{}\footnote{#1}%
  \addtocounter{footnote}{-1}%
  \endgroup
}

\usepackage{xcolor}
\usepackage{colortbl}
\usepackage{makecell}

\definecolor{azure}{rgb}{0.21, 0.49, 0.92}

\newcommand{\physics}{\textit{physics}}
\newcommand{\semantic}{\textit{semantic}}
\newcommand{\intrinsic}{\textit{intrinsic}}
 
\title{\Large{Toulouse Hyperspectral Data Set:} \\ \large{A benchmark data set to assess semi-supervised spectral \\ representation learning and pixel-wise classification techniques}}

\author[1,$\ast$]{Romain Thoreau}
\author[4]{Laurent Risser}
\author[2]{Véronique Achard}
\author[3]{Béatrice Berthelot}
\author[2]{Xavier Briottet}

\affil[1]{CNES, FR-31401 Toulouse, France}
\affil[2]{ONERA-DOTA, University of Toulouse, FR-31055
Toulouse, France}
\affil[3]{Magellium, 31520 Ramonville Saint-Agne, France}
\affil[4]{Toulouse Mathematics Institute (UMR 5219), CNRS,
University of Toulouse, F-31062 Toulouse, France}

\begin{document}
\maketitle

\begin{abstract}
Airborne hyperspectral images can be used to map the land cover in large urban areas, thanks to their very high spatial and spectral resolutions on a wide spectral domain. 
While the spectral dimension of hyperspectral images is highly informative of the chemical composition of the land surface, the use of state-of-the-art machine learning algorithms to map the land cover has been dramatically limited by the availability of training data.
To cope with the scarcity of annotations, semi-supervised and self-supervised techniques have lately raised a lot of interest in the community.
Yet, the publicly available hyperspectral data sets commonly used to benchmark machine learning models are not totally suited to evaluate their generalization performances due to one or several of the following properties: a limited geographical coverage (which does not reflect the spectral diversity in  metropolitan areas), a small number of land cover classes and a lack of appropriate standard train / test splits for semi-supervised and self-supervised learning. 
Therefore, we release in this paper the Toulouse Hyperspectral Data Set that stands out from other data sets in the above-mentioned respects in order to meet key issues in spectral representation learning and classification over large-scale hyperspectral images with very few labeled pixels. Besides, we discuss and experiment self-supervised techniques for spectral representation learning, including the Masked Autoencoder \cite{he2022masked}, and establish a baseline for pixel-wise classification achieving 85\% overall accuracy and 77\% F1 score. The Toulouse Hyperspectral Data Set and our code are publicly available at \href{https://www.toulouse-hyperspectral-data-set.com}{www.toulouse-hyperspectral-data-set.com} and \href{https://www.github.com/Romain3Ch216/tlse-experiments/}{www.github.com/Romain3Ch216/tlse-experiments/}, respectively.
\end{abstract}

\blfootnote{$^\ast$Corresponding author: romain.thoreau@cnes.fr}
\blfootnote{Work done while at ONERA / Magellium}

\section{Introduction \label{sec:intro}}

Airborne hyperspectral images are a critical resource for the land cover mapping of large urban areas. As much as artificial impermeable surfaces impact watershed hydrology (particularly droughts and floods) \cite{labbas, bras, desbordes, dosdogru, giri}, urban heat island effects \cite{onishi, ipcc2022} and soil carbon uptake \cite{pereira2021urban, soil-7-661-2021, eu2012guidelines, scalenghe2009anthropogenic}, providing public authorities and scientists with accurate maps of land surface materials is a key issue to mitigate the effects of urban sprawl. Hyperspectral sensors measure the radiant flux reflected by the ground and by the atmosphere for several hundreds of narrow and contiguous spectral intervals in the visible and the reflective part of the infrared. While the spectral radiance measured at the sensor level partially depends on the atmosphere (\textit{i.e.} its water vapor concentration, type and concentration of aerosols, etc.), atmospheric correction algorithms such as \cite{miesch2005direct, gao2009atmospheric} can estimate the pixel-wise reflectance at the ground level, which is the ratio of the reflected radiant flux on the incident radiant flux averaged over the pixel surface. Reflectance is intrinsic to the chemical composition of matter, and is therefore very informative of the land cover. In contrast, the spatial information brings little information (for instance, an orange tennis court could actually either be in porous concrete or in a synthetic track, which would be indistinguishable with a conventional RGB image) though a large-scale context may sometimes raise ambiguities. 

The main hindrance to the pixel-wise classification of hyperspectral images holds in the scarcity of labeled data. 
Labeling pixels of a hyperspectral image with land cover classes of low abstraction such as \textit{gravel} or \textit{asphalt} indeed requires expert knowledge and expensive field campaigns. 
Therefore, the ground truth that usually contains at most 1\% of the pixels barely represents the spectral variability of the image. 

To that extent, publicly available hyperspectral data sets have fueled a great deal of research in several directions including Active Learning \cite{tuia_review, 9774342}, unsupervised / self-supervised and semi-supervised learning \cite{camps2007semi, wu2017semi, sawant2020review, yue2021self}, to train machine learning models with few labeled samples that are robust to spectral intra-class variations. Nevertheless, if the Standardized Remote Sensing Data Website\footnote{\url{http://dase.grss-ieee.org/index.php}} of the IEEE Geoscience and Remote Sensing Society (GRSS) provides a set of community data sets and a tool to evaluate classifiers on undisclosed test samples, providing the ground truth of public data sets with standard training sets (divided in a subset for the supervised part and another subset for the unsupervised part) spatially disjoint to test sets would foster reproducible and fair evaluations of semi-supervised techniques. We emphasize that several works including \cite{audebert2019deep, lange2018influence, geiss2017effect} showed that random sampling of the training and test sets over-estimates the generalization performances of classifiers, which is partly explained by the fact that pixels belonging to the same semantic class but sampled in different geographical areas are obviously more likely to have different spectral signatures than neighboring pixels. 

Therefore, we introduce the Toulouse Hyperspectral Data Set\footnote{The Toulouse Hyperspectral Data Set is available at \href{https://www.toulouse-hyperspectral-data-set.com}{www.toulouse-hyperspectral-data-set.com}} that better represents the complexity of the land cover in large urban areas compared to currently public data sets, and provide standard training and test sets specifically defined to assess semi/self-supervised representation learning and pixel-wise classification techniques. First, we present the construction and the properties of the Toulouse data set in section \ref{sec:properties}. Second, we provide a qualitative comparison with the Pavia University\footnote{\url{https://www.ehu.eus/ccwintco/index.php/Hyperspectral_Remote_Sensing_Scenes}} and Houston University \cite{houston} data sets in section \ref{sec:comparison}. 
Third, we discuss and experiment\footnote{Code to reproduce our experiments is available at \href{https://www.github.com/Romain3Ch216/tlse-experiments/}{www.github.com/Romain3Ch216/tlse-experiments/}} self-supervised techniques for pixel-wise classification in section \ref{sec:self-supervision}. Finally, we conclude in section \ref{sec:conclusion}.

\section{Construction and properties of the Toulouse Hyperspectral Data Set \label{sec:properties}}

\begin{figure}
	\center
	\includegraphics[width=\textwidth]{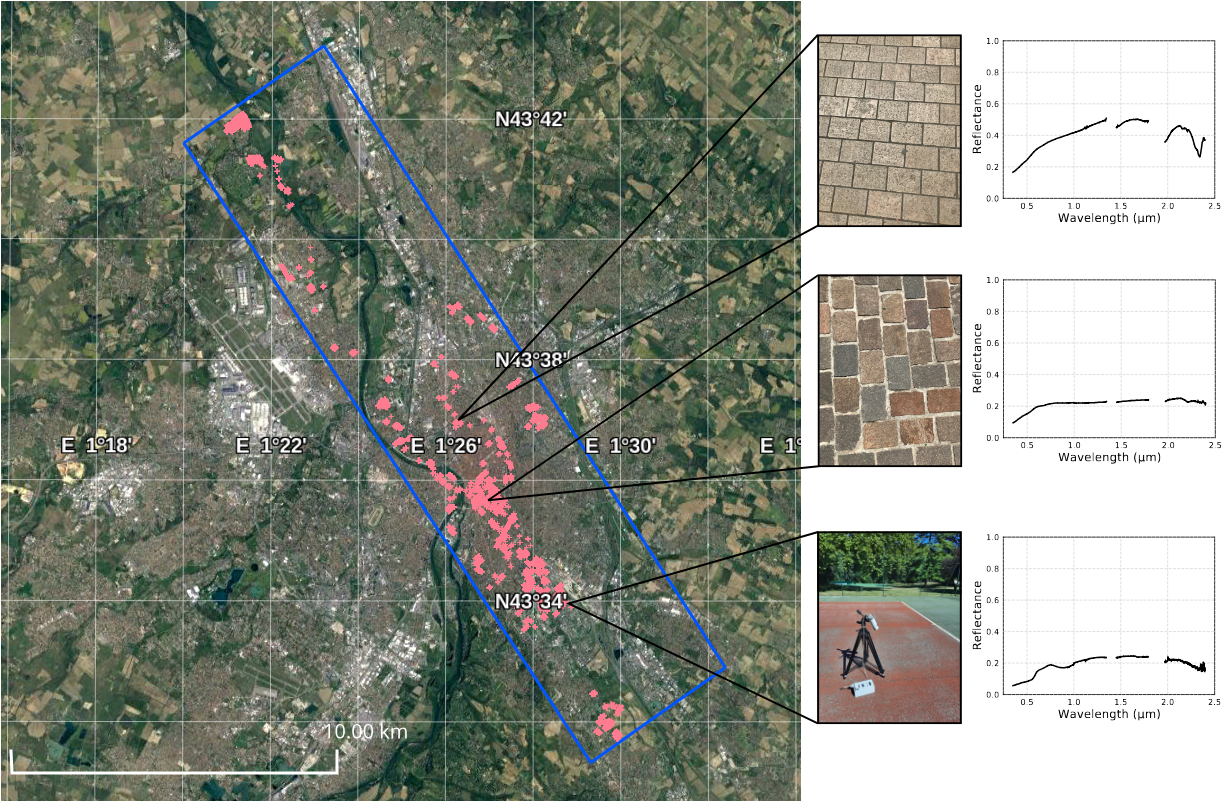}
	\caption[Area of Toulouse covered by the AI4GEO airborne hyperspectral image]{Area of Toulouse covered by the AI4GEO airborne hyperspectral image (in blue), our annotated ground truth (in red), and examples of reflectance spectra (clear paving stone, brown paving stone and red porous concrete, from top to bottom) measured on field with ASD spectrometers during the CAMCATT-AI4GEO field campaign \cite{ROUPIOZ2023109109}. \label{fig:toulouse_area}}
\end{figure}

In the context of the AI4GEO consortium\footnote{\url{https://www.ai4geo.eu/en}} and the CAMCATT/AI4GEO field campaign \cite{ROUPIOZ2023109109}, a hyperspectral image was acquired over the city of Toulouse the $15^{th}$ of June 2021 around 11am UTC with a AisaFENIX 1K camera (which has a spectral range from 0.4 µm to 2.5 µm with a 3.6 nm spectral resolution in the VNIR\footnote{Visible and near infrared} and a 7.8 nm specral resolution in the SWIR\footnote{Short-wave infrared}, a swath of 1024 m and a ground sampling distance of 1 m) that was on-board a Safire  aircraft that flew at 1,500 m above the ground level. The hyperspectral data was converted in radiance at aircraft level through radiometric and geometric corrections. Then, the radiance image was converted to surface reflectance with the atmospheric correction algorithm COCHISE \cite{miesch2005direct}. Hyperspectral surface reflectances were also acquired on-ground with three ASD spectrometers in the range of 0.4 µm to 2.5 µm. Reflectance spectra of \textit{clear paving stone}, \textit{brown paving stone} and \textit{red porous concrete} with pictures of the materials are shown in Fig. \ref{fig:toulouse_area} as examples. These in-situ measurements have served as a basis to define a land cover nomenclature (several materials with in-situ measurements are not in our nomenclature because they were on walls or on small manhole covers for instance) and to build a ground truth by photo-interpretation, additional field campaigns as well as with the help of exogenous data. Precisely, we used the "Registre Parcellaire Graphique"\footnote{\url{https://artificialisation.developpement-durable.gouv.fr/bases-donnees/registre-parcellaire-graphique}}, a geographical information system that informs the crop type of agricultural plots over France, to annotate cultivated fields. 
For a full description of the data acquired in the CAMCATT / AI4GEO campaign, we refer the reader to the data paper \cite{ROUPIOZ2023109109}. 

\begin{figure}
	\includegraphics[width=1\textwidth]{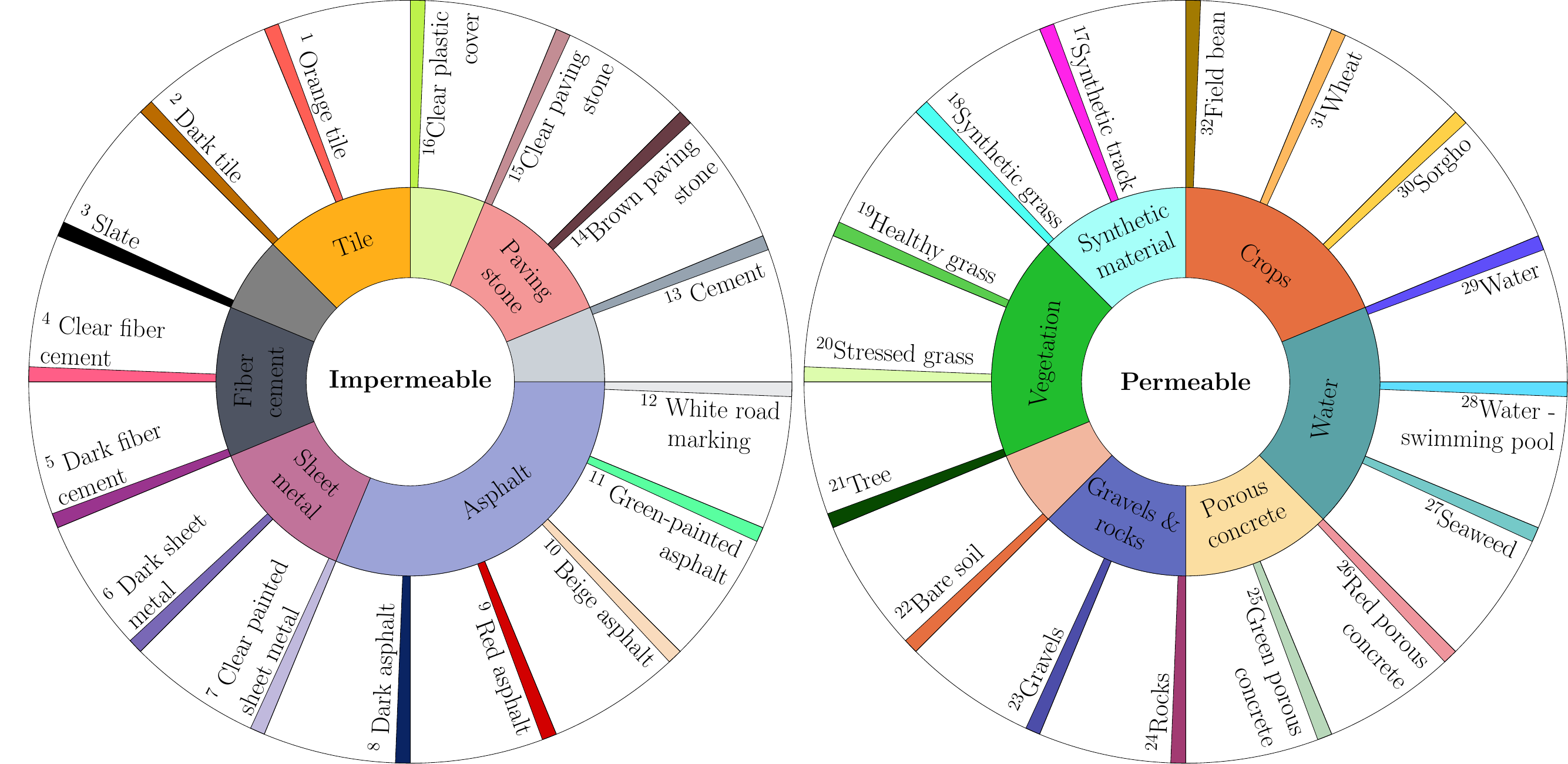}
	\caption{Land cover nomenclature of Toulouse Hyperspectral Data Set \label{fig:nomenclature}}
\end{figure}

\subsection{Land cover ground truth}

In total, we define the land cover nomenclature with 32 classes, dividing into 16 impermeable materials and 16 permeable materials, that we organize in a hierarchical nomenclature as shown in Fig. \ref{fig:nomenclature}. Approximately $380,000$ pixels are labeled with a land cover class. In contrast to conventional semantic segmentation data sets, our ground truth is made of sparse annotations, \textit{i.e.} polygons that are disconnected from each other. We annotated the pixels with particular attention to the exactness of the land cover labels. In particular, we omitted from the ground truth, as much as possible, mixed pixels (\textit{i.e.} pixels with several materials) whose reflectance spectra are a combination of various spectra. Random spectra of the classes \textit{orange tile} and \textit{synthetic track} are shown in Fig. \ref{fig:two-spectra} while the other classes are shown in the appendices in Fig. \ref{fig:spectra}.

\subsection{Standard training and test sets for semi-supervised learning \label{sec:split}}

We provide 8 spatially disjoint splits of the ground truth divided into:
\begin{itemize}
	\item A labeled training set,
	\item An unlabeled training set divided itself in:
	\begin{itemize}
		\item A set exclusively composed of pixels belonging to land cover classes defined in the nomenclature, called the \textit{labeled pool},
		\item A larger set of truly unlabeled pixels that may not belong to known land cover classes, called the \textit{unlabeled pool},
	\end{itemize}
	\item A validation set,
	\item A test set.
\end{itemize}

As much as the spectral intra-class variability of hyperspectral data comes from 1) variations in illumination conditions, 2) non-lambertian effects and slight variations in the material chemical composition (for instance due to variations in water or chlorophyll content of different trees from the same specie, variations of tar due to aging or to different environmental exposures) and 3) from larger variations in the material composition, due to the fact that the nomenclature gathers different materials under the same class (for instance different tree species gathered in a unique class) \cite{revel2018inertia}, there are high correlations between the intra-class spectral variability and the geographical location of pixels. Therefore, we suggest to foster the statistical independence of the training, validation and test sets by spatially separating them (see an example in Fig. \ref{fig:polygons} in the appendices). 

To perform the spatially disjoint splits of the ground truth while ensuring that each class is distributed in appropriate proportions in the labeled training set, the validation set and the test set, we group neighboring polygons together in $n_{groups}$ groups and define the ground truth split as mixed integer problem. Precisely, we aim to assign to each group of neighboring polygons a set, among the labeled training set (designated by index 1), the labeled pool (designated by index 2), the validation set (designated by index 3) and the test set (designated by index 4), while the unlabeled pool is left out. The main constraint of the problem is that each set $s \in \{1, 3, 4\}$ should contain, for each class, at least $p_s$ percent of the total number of labeled pixels.  We define the mixed integer problem as follows where $u_{ij}$ is 1 if group $i$ is in set $j$, $P[i,k]$ is the number of pixels of class $k$ in group $i$ and $\mathcal{S} = \{1, 3, 4\}$:

\begin{align}
	\min_{u} & \sum_{s \in \mathcal{S}} \sum_{i=1}^{n_{groups}} \sum_{k=1}^c P[i,k] \cdot u_{is} & \\
	\textrm{subject to:} & \sum_{j=1}^4 u_{ij} = 1 \:\:\:\: \mbox{\textit{i.e. each group should be at least in one set}} \\
	& \forall s \in \mathcal{S}, \forall k \in \{1, \ldots, c\}, \sum_{i=1}^{n_{groups}} P[i,k] \cdot u_{is} \geq p_s \cdot \sum_{i=1}^{n_{groups}} P[i,k] \\
	& \mbox{\textit{i.e. for each class k, the proportion of pixels in set }} s \nonumber \\ 
	& \mbox{\textit{ should be greater than the proportion }} p_s \nonumber
\end{align}

In the standard splits that we provide, 13\%, 29\%, 14\% and 46\% of the labeled samples are in the labeled training set, the labeled pool, the validation set and the test set, respectively, in average (with regard to classes and to the 8 splits). In addition, the unlabeled pool contains nearly 2.6 million pixels. Hence, the labeled pixels used for training only represent 7\% of all data. The decision to divide the ground truth into splits of 13\%, 29\%, 14\% and 46\% of the total number of labeled pixels stems from the following considerations, in order of priority: having a representative test set, having a representative validation set, having a sufficient number of samples in the training set for supervision to be relevant. However, the precise choice of the average proportions in each set is arbitrary and does not rely on a statistical analysis. In addition, we chose to provide only height splits of the ground truth because we could not find other solutions of the mixed integer problem that were significantly different from each other.

\begin{figure}
	\center
	\begin{subfigure}{0.45\textwidth}
		\includegraphics[width=\textwidth]{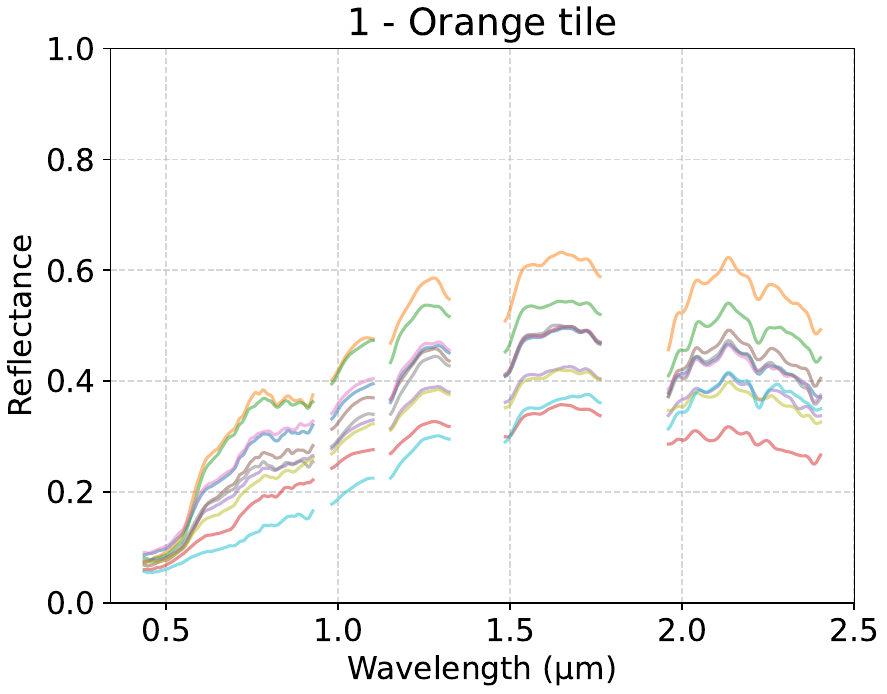}
	\end{subfigure}
	\hfill
	\begin{subfigure}{0.45\textwidth}
		\includegraphics[width=\textwidth]{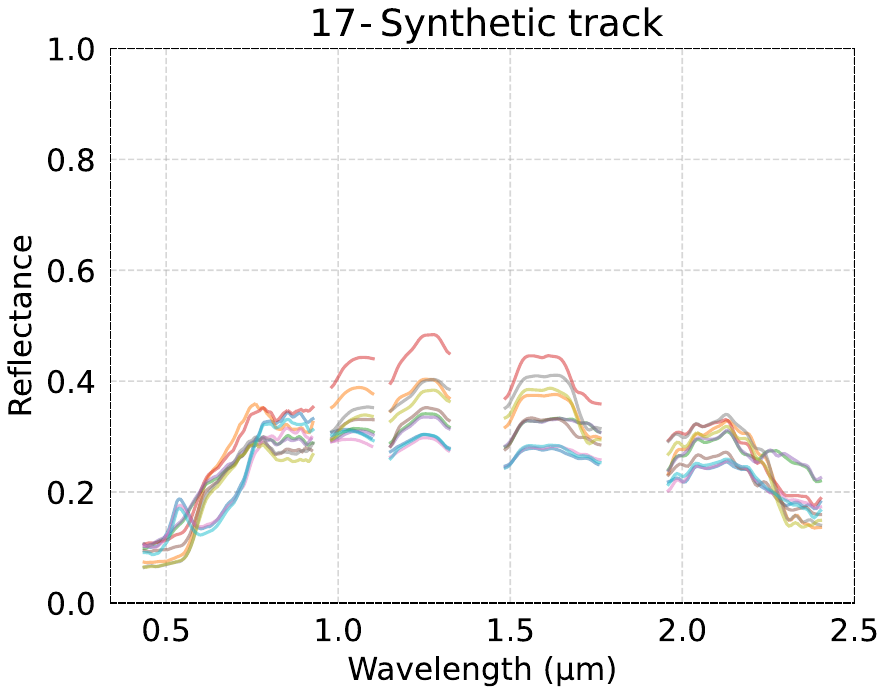}
	\end{subfigure}
	\caption{Random spectra of the classes \textit{orange tile} and \textit{synthetic track}. \label{fig:two-spectra}}
\end{figure}

\subsection{Python package}

Since the hyperspectral images of the Toulouse data set are too large to be loaded into memory all at once, we release \href{https://github.com/Romain3Ch216/TlseHypDataSet}{TlseHypDataSet}, a Python library whose main objective is to enable easy and rapid loading of the data into Pytorch\footnote{https://pytorch.org/docs/stable/index.html} loaders.

\begin{figure}[h]
\begin{python}
import torch
from TlseHypDataSet.tlse_hyp_data_set import TlseHypDataSet
from TlseHypDataSet.utils.dataset import DisjointDataSplit

dataset = TlseHypDataSet('/path/to/dataset/', patch_size=1)

# Load the first standard ground truth split
ground_truth_split = DisjointDataSplit(dataset, split=dataset.standard_splits[0])

train_loader = torch.utils.data.DataLoader(ground_truth_split.sets_['train'], shuffle=True, batch_size=1024)

for epoch in range(100):
    for samples, labels in train_loader:
        ...
\end{python}
\caption{Minimal example of Python code to load data in Pytorch loaders with the \href{https://github.com/Romain3Ch216/TlseHypDataSet}{TlseHypDataSet} library}
\end{figure}

\section{Comparison with publicly available data sets \label{sec:comparison}}

In this section, we compare different properties of the Toulouse data set that are meaningful for machine learning applications, to those of the Pavia University data set and the Houston University data set. These two data sets are indeed widely used in the community, and cover urban or peri-urban areas as well. 

\textbf{Spectral and spatial information} Tab. \ref{tab:resolutions} recaps the spatial and spectral resolutions as well as the spectral domains of the compared data sets. While the spatial resolution is roughly the same, the Toulouse image has a much wider spectral domain which significantly brings more discriminating information for the mapping of the land cover, especially for mineral materials which are numerous in urban areas.

\begin{figure}
	\includegraphics[width=0.9\textwidth]{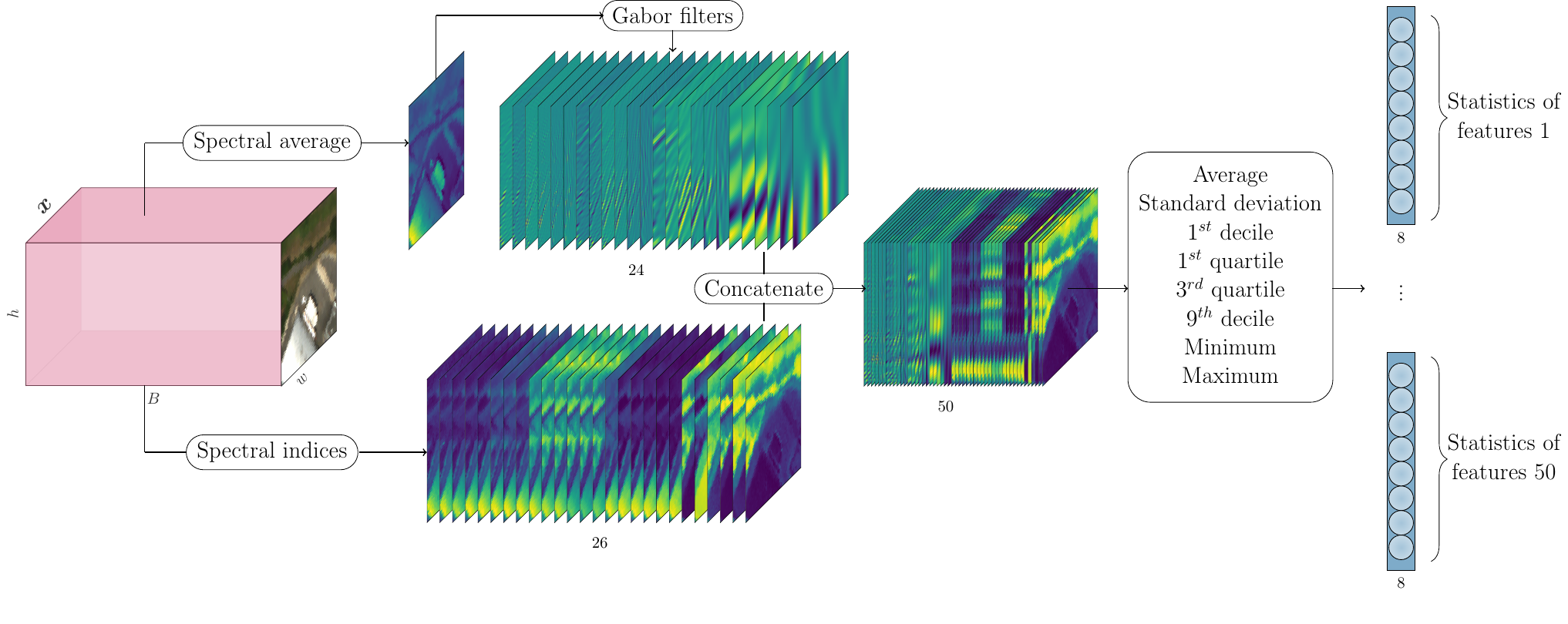}
	\caption[Illustration of our hand-crafted patch-wise feature extraction technique]{Illustration of our hand-crafted patch-wise feature extraction technique. The input is a 64 by 64 pixel hyperspectral patch. On one side, spectral indices (which include a selection of 20 spectral bands uniformly sampled along the spectral domain) are computed, resulting in 26 maps of 64 by 64 pixels. On the other side, the patch, averaged along the spectral dimension, is filtered by Gabor filters with 4 different frequencies (from 1 m$^{-1}$ to 10 m$^{-1}$) and 6 different orientations, resulting in 24 maps. From every maps, spatial statistics are computed, resulting in a 400-dimensional feature.  \label{fig:features}}
\end{figure}

\textbf{Spectral and spatial variability} Comparing data sets with different spatial and spectral resolutions as well as different spectral domains is not straightforward. If the resolutions and spectral domains were the same, we could consider to learn representations of hyperspectral patches with an autoencoder and visualize the representations with a 2-dimensional t-SNE \cite{van2008visualizing} transformation to qualitatively compare the data sets as in \cite{castillo2021semi}. Instead, we suggest to represent $64\times64$ pixel hyperspectral patches (that are, at least, partially labeled) with hand-crafted features. To summarize the spectral information of patches, we compute spectral indices (\textit{i.e.} linear combinations of spectral bands, sometimes normalized) that use spectral channels included in the smallest spectral domain of the compared data sets (here Pavia University), precisely the NDVI \cite{sellers1985canopy}, ANVI \cite{pena2007mapping}, CI \cite{bausch2010quickbird}, NDVI\_RE \cite{gitelson1994spectral}, VgNIR\_BI \cite{estoque2015classification}, SAVI \cite{huete1988soil}, and uniformly sample 20 bands over the whole domain. To summarize the spatial information of patches, we compute 24 predefined Gabor filters on the spectral average of the patch (4 different frequencies (from 1 m$^{-1}$ to 10 m$^{-1}$) and 6 different orientations). Then, we concatenate the spectral  and spatial features and compute patch-wise statistics (the average, the standard deviation, the first and last deciles, the first and last quartiles, as well as the minimum and the maximum), yielding a 400-dimensional feature for each patch. The representation method is illustrated in Fig. \ref{fig:features}.

\begin{figure}
	\center
	\includegraphics[width=0.7\textwidth]{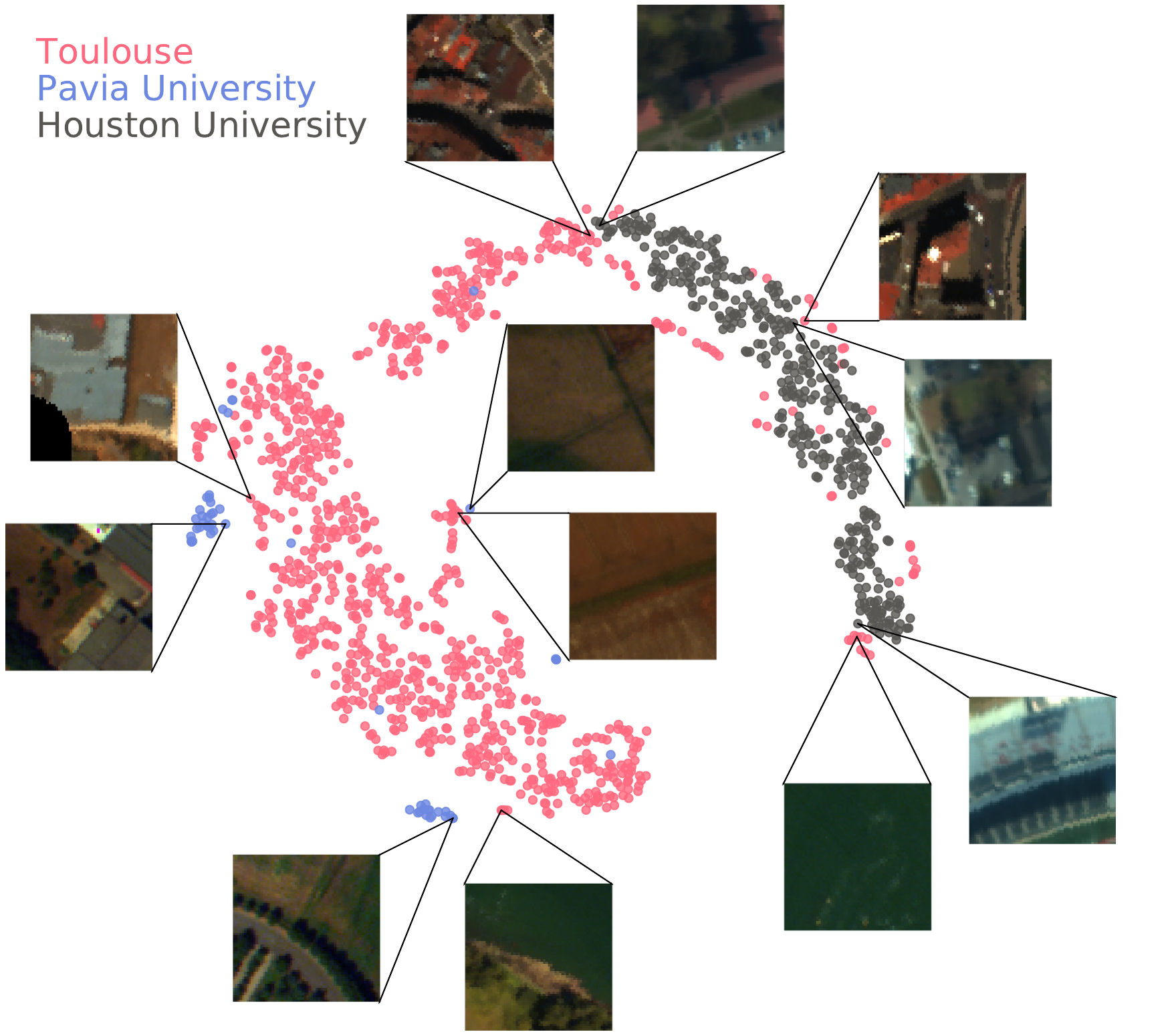}
	
	\caption{t-SNE visualization of hand-crafted representations of $64\times64$ pixel hyperspectral patches from the Pavia University, Houston University and Toulouse data sets.\label{fig:proj}}
\end{figure}

Fig. \ref{fig:proj} shows a t-SNE visualization of the hand-crafted features. First, the Toulouse data set clearly occupies more space than the Pavia and Houston data sets. Second, it seems that similar landscapes are projected in the same regions, as show a few false-color composition of hyperspectral patches close in the 2-dimensional space but taken from different data sets. In order to compare the contribution of the spatial information, spectral information and spectral information in the SWIR\footnote{short-wave infrared} (the spectral domains of Pavia and Houston are limited to the NIR\footnote{near infrared}), we made additional comparisons presented in the appendices in Fig. \ref{fig:proj2}.  
We found that the larger variability of Toulouse mainly comes from the spectral dimension and is mainly a consequence of a larger variability of the land cover.

\begin{figure}[h]
	\center
	\begin{subfigure}{0.32\textwidth}
		\center
		\includegraphics[width=0.9\textwidth]{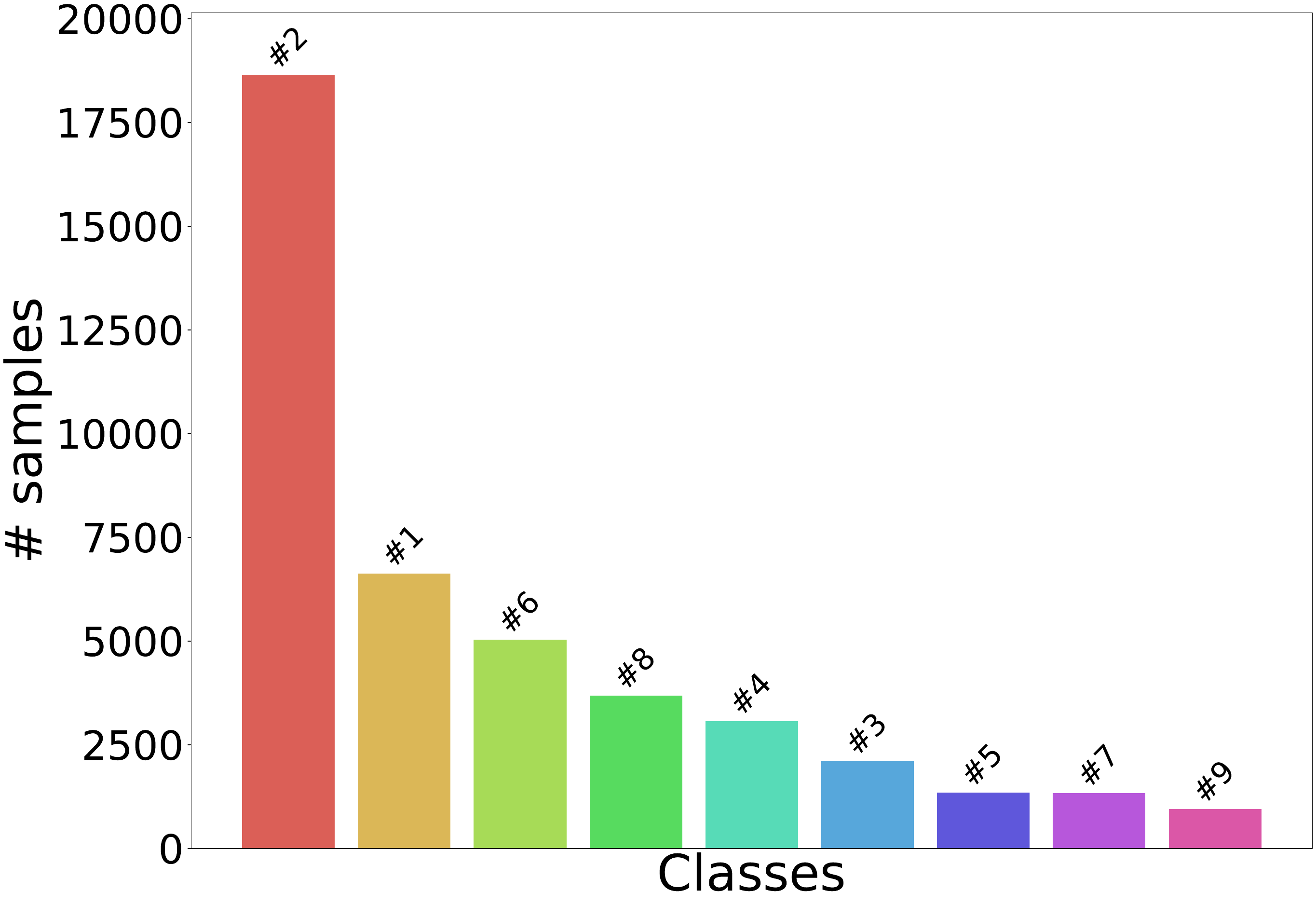}
		\caption{Pavia University}
	\end{subfigure}
	\begin{subfigure}{0.32\textwidth}
		\center
		\includegraphics[width=0.9\textwidth]{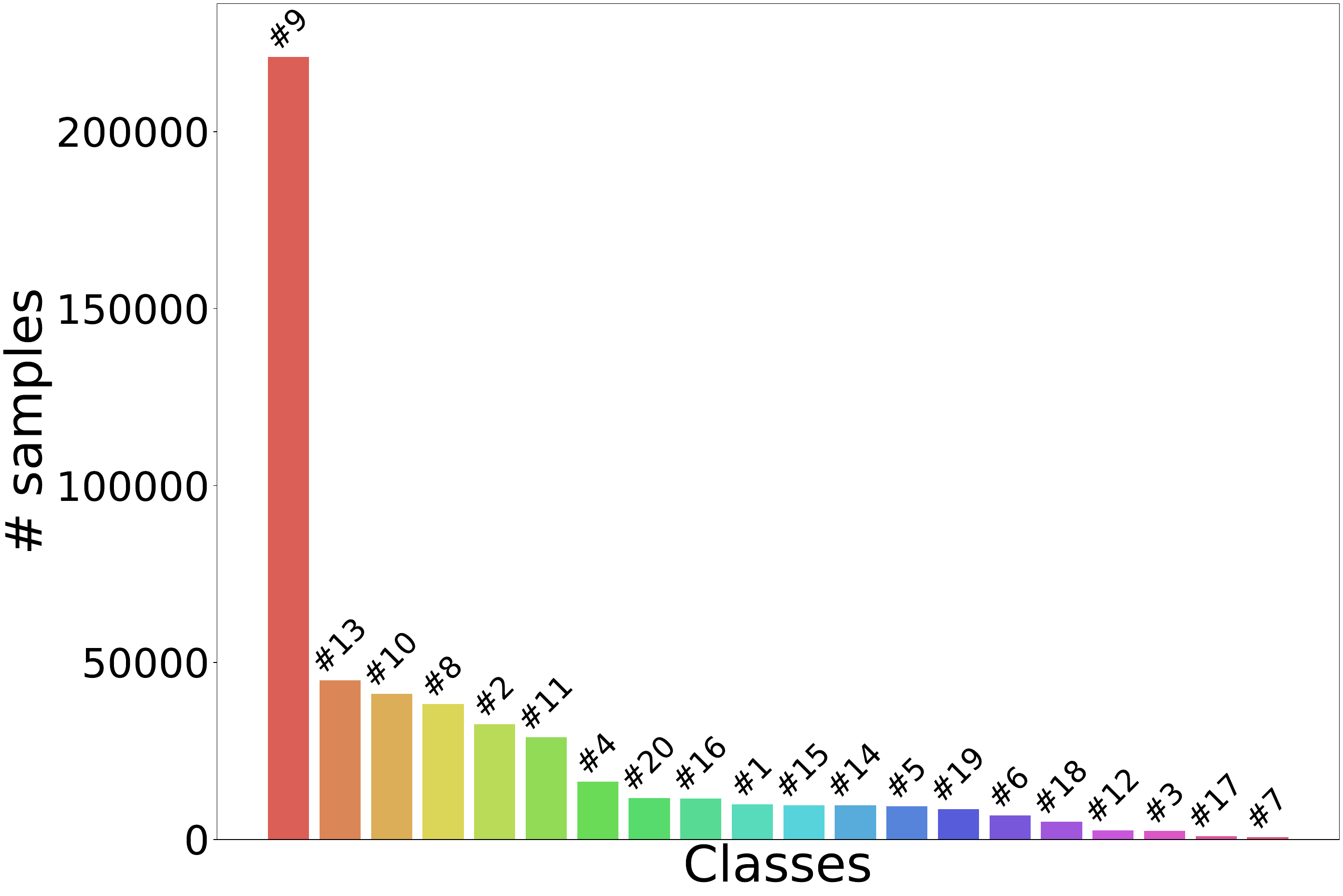}
		\caption{Houston university}
	\end{subfigure}
	\begin{subfigure}{0.32\textwidth}
		\center
		\includegraphics[width=0.9\textwidth]{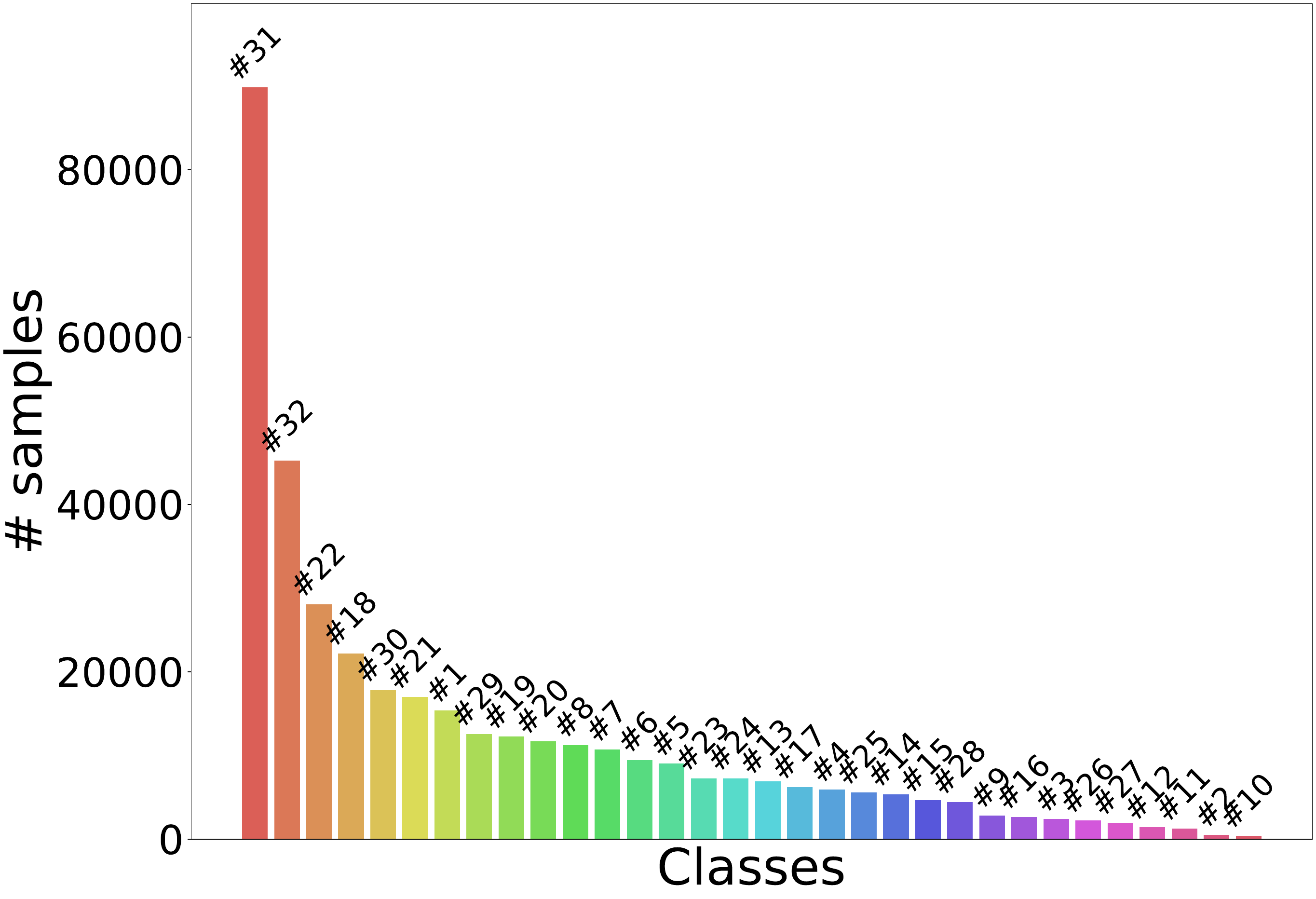}
		\caption{Toulouse}
	\end{subfigure}
	\caption{Number of samples by class sorted from the most to the less represented. \label{fig:long-tailed}}
\end{figure}

\textbf{Class distribution} Tab. \ref{tab:stats} recaps statistics about the class distribution of the data sets, including the imbalance ratio, which is the ratio of the number of samples in the largest class over the number of samples in the smallest class. The Houston and Toulouse data sets are particularly imbalanced: the biggest class of Houston accounts for 43\% of the samples while the biggest class of Toulouse accounts for 24\% of the samples. Yet, Fig. \ref{fig:long-tailed} shows that the Toulouse data set particularly exhibits a long-tailed class distribution, which is representative of life-like scenarios. Data sets with a long-tailed class distribution are data sets where a small number of classes account for a large part of samples while a large number of classes have only few examples \cite{zhang2023deep}. The difference between usual class imbalance and long-tailed class imbalance mainly lies in the number of classes with few samples. 

\textbf{Noisy labels} Compared to the Houston data set, we argue that the Toulouse data set contains less noise in the ground truth. Although the Houston data set contains more than a half million labeled pixels, many pixels are wrongly labeled, or are at least misleading as there are a mix of several materials. This noise in the ground truth is detrimental to classification models that put more emphasize on the spectral information rather than the spatial information. We show in the appendices a few examples of noisy labels in Fig. \ref{fig:noise} and illustrate in Fig. \ref{fig:examples_tlse} the care we took to avoid noisy labels in the Toulouse ground truth. 


\begin{table}[h]
	\caption{Spectral and spatial characteristics of the Pavia, Houston and Toulouse hyperspectral data sets. \label{tab:resolutions}}
	\begin{center}
	\begin{tabular}{lcccc}
		\toprule
		\textbf{Data set} & \textbf{GSD} & \textbf{Spectral domain} & \textbf{Spectral resolution} & \textbf{Spectral dimension} \\
		Pavia University & 1.3 m & 0.43 µm - 0.86 µm & 4 nm & 103\\
		Houston University & 1 m & 0.38 µm - 1.0 µm & 3.5 nm & 48\\
		Toulouse & 1 m & 0.4 µm - 2.5 µm & 3.6 nm (VNIR) \& 7.8 nm (SWIR) & 310 \\
	\end{tabular}
	\end{center} 
\end{table}

\begin{table}[h]
	\caption{Statistics of the Pavia, Houston and Toulouse hyperspectral data sets. \label{tab:stats}}
	\begin{center}
	\begin{tabular}{lccc}
		\toprule
		\textbf{Data set} & \textbf{\# classes} & \textbf{\# data samples} & \textbf{Imbalance ratio} \\
		Pavia University & 9 & 43,000 & 20 \\
		Houston University & 20 & 510,000 & 329 \\
		Toulouse & 32 & $380,000$ & 225 \\
	\end{tabular}
	\end{center} 
\end{table}

%
%
%
%
%

\clearpage
\section{Self-supervision for spectral representation learning \label{sec:self-supervision}}

In this section, we review state-of-the-art self-supervised techniques, discuss their applicability to hyperspectral data and establish a first semi-supervised baseline on the Toulouse Data Set that shall serve as comparison for future works.

\subsection{Self-supervised learning: an overview}

Common semi-supervised learning techniques jointly optimize machine learning models on a supervised task and on an auxiliary unsupervised task. A common choice for the auxiliary task is the reconstruction of the (high dimensional) data from a (low dimensional) representation. However, a wide range of new approaches, known as self-supervised learning techniques, have recently emerged by introducing more useful auxiliary tasks. Self-supervision consists in training the model on a supervised pretext task for which labels are automatically generated. 

In computer vision, a variety of pretext tasks have emerged in order to learn similar visual representations for different views of the same data. Those tasks include rotation self-supervision \cite{gidaris2018unsupervised}, exemplar self-supervision \cite{dosovitskiy2014discriminative}, contrastive learning \cite{wu2018unsupervised, henaff2020data, oord2018representation, tian2020contrastive, chen2020simple} or self-supervised knowledge distillation \cite{caron2021emerging}. Rotation self-supervision aims to predict the rotation (0°, 90°, 180° or 270°) applied to an image. Exemplar self-supervision gathers transformations of the same data sample under one class and trains the model on the subsequent classification task. Contrastive learning consists in learning similar representations of automatically-generated pairs of data samples with common semantic properties (positive pairs), while learning dissimilar representations of unrelated data samples (negative pairs). In particular, the seminal framework of \cite{chen2020simple} is based on stochastic data augmentation (specifically the combination of random cropping, random color distortions and random Gaussian blur) and a contrastive loss function that aims to identify a positive pair within a batch of samples. Self-supervised knowledge distillation takes inspiration from knowledge distillation \cite{hinton2015distilling} by taking a teacher network to supervise a student network that sees different transformations of the input data. Self-supervised learning has been integrated in semi-supervised learning frameworks, for instance in \cite{zhai2019s4l} or \cite{fini2023semi} that combine cluster-based self-supervision with class prototype learning.

To summarize, these strategies of self-supervision generate different views of the data (far apart in the input space) with the same semantic content thanks to data augmentation techniques, and rely on various tricks to prevent representation collapse (when the encoder learns useless representations that nevertheless minimize the training objective) \cite{assran2023self}. They have experimentally demonstrated benefits on RGB natural images in term of robustness to various spatial contexts (\textit{e.g.} pose, orientation, background...), which are the main cause of intra-class variability (\textit{e.g.} a car in a parking lot and the same car on a freeway). 
In contrast, the spectral intra-class variability of hyperspectral data does not depend on context and divides into \physics, \intrinsic \ and \semantic \ intra-class variabilities, as discussed in section \ref{sec:split}. 

Many self-supervised learning techniques have been directly applied to hyperspectral data, such as \cite{9734031} that augment hyperspectral patches with random cropping and random color distortions, and \cite{duan2022self} that apply random rotations as well as spectral random noise and spectral mirroring, both in the framework of self-supervised contrastive learning. Here we shall note that the physical soundness of random color distortions and spectral mirroring should be questioned, and that those augmentations are unlikely to preserve the semantic information of the data. Other works have introduced data augmentation techniques that are specific to hyperspectral data, such as \cite{9658507} that creates positive pairs of data samples by sampling monochromatic images from neighboring spectral channels, or \cite{rs15061713} that pairs spectrally close samples. All in all, most attention has been put on learning spatial-spectral representations with self-supervised techniques, often based on data augmentation, though we believe that finding a data augmentation technique that is faithful to realistic \physics, \intrinsic \ and \semantic \ intra-class variations is not trivial, if not impossible. As a matter of fact, the true illumination conditions (that depend on topography) should be known to simulate realistic \physics \ variations, while \intrinsic \ and \semantic \ spectral variations are, by nature, intrinsic to the chemical composition of matter.

\begin{figure}
	\center
	\includegraphics[width=0.8\textwidth]{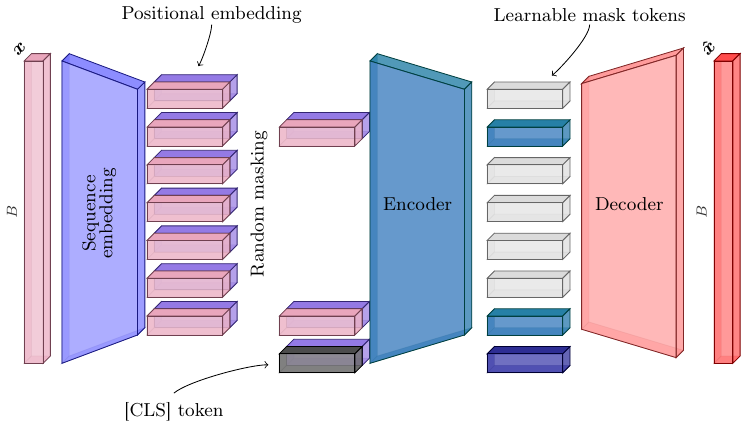}
	\caption{Illustration of the Masked Autoencoders \cite{he2022masked} for 1D data. The $1 \times B$ dimensional input spectrum $\boldsymbol{x}$ is divided in small sequences. A large part of the sequences are randomly masked. The visible spectral sequences (with positional encoding) are encoded by the transformer. Then, learnable mask tokens are concatenated with the $1 \times d$ dimensional embeddings of the sequences, that are mapped to the reconstructed data $\hat{\boldsymbol{x}}$ by the light-weight decoder.  \label{fig:mae}}
\end{figure}

  
Therefore, we focus on two self-supervised techniques that do not rely on data augmentation: Deep Clustering \cite{caron2018deep} and Masked Autoencoders (MAE) \cite{he2022masked}. \\
Deep Clustering is a seminal technique in cluster-based self-supervision that uses pseudo-labels derived from a clustering algorithm to supervise the training. The core idea behind Deep Clustering is that the use of convolutional layers for learning visual representations is a very strong inductive bias about the data structure. As a matter of fact, a dense network fine-tuned on top of the frozen features computed by a randomly initialized CNN, namely AlexNet, achieved 12\% accuracy on ImageNet which is far above chance (\textit{i.e.} the performance of a classifier with a uniform predictive distribution) \cite{noroozi2016unsupervised}. Deep Clustering leverages this prior on the input signal to iteratively learn representations from the supervision of pseudo-labels obtained by a standard clustering algorithm performed in the feature space. \\
Concerning MAE, it strongly masks the input data and learns to reconstruct its missing parts. For reflectance spectra, as much as the combination of spectral features (absorption peaks, spectral inflexion, etc.) at different wavelengths is closely related to the chemical composition of the land surface,  the \textit{masked reconstruction} task of MAE seems particularly relevant to learn discriminating features (of the materials) without class supervision. While MAE were used on $8 \times 8$ pixel spaceborne hyperspectral patches (> 10 m GSD) in \cite{zhu2023spectralmae}, we are interested in this paper on spectral representations only as the spectral information is much more discriminating of the materials on the ground surface than the spatial information in the case of airborne hyperspectral images. The Masked Autoencoder for 1D data is illustrated in Fig. \ref{fig:mae}. Originally, MAE processes RGB images that are divided in small patches which are encoded and decoded by vision transformers \cite{dosovitskiy2021image}. While transformers \cite{vaswani2017attention} have been adapted for hyperspectral data in \cite{hong2021spectralformer}, we use a simpler architecture and make as few changes as possible from the original MAE, keeping in mind two important points: 1) in contrast to words that are very abstract concepts (or to small image patches that can contain high-level information), reflectance values are not meaningful by themselves, 2) in contrast to words or to image patches, the relative distance between spectral channels does not contain semantic information. By this we mean that the position of a word in a sentence contains valuable semantic information, while the distance between two spectral features simultaneously observed on a spectrum is not informative\footnote{For instance, a random permutation of the spectral channels would not decrease the performance of dense neural networks on a classification task}. This is why we believe that the transformer architecture is not particularly relevant for hyperspectral data, but is very convenient for the masked reconstruction task.

\subsection{Experiments on the Toulouse Hyperspectral Data Set}

\subsubsection{Experimental protocol}

In this section, we experimentally evaluated the potential of the spectral representations learned by Deep Clustering and MAE for a downstream classification task.

We compared the representation learning potential of the MAE against a conventional autoencoder in terms of classification accuracy. We trained a k-nearest neighbor algorithm (KNN) and a Random Forest classifier (RF) applied on the latent space of the autoencoder, on the [CLS] tokens of the masked autoencoder, and on the raw data as a baseline.

The rationale behind the use of the KNN classifier is that a self-supervised pretrained model should output grouped representations of semantically similar data \cite{reed2023scale}. Recent and prominent works have used the KNN classifier to evaluate the discriminating potential of representations learned by self-supervised models \cite{caron2021emerging, chen2021exploring, wu2018unsupervised}.
Other common validation protocols are to learn a linear classifier on top of the frozen encoder or to fine-tune the self-supervised model on the classifcation task \cite{caron2021emerging, he2020momentum}. We found that a RF classifier on top of the frozen features outperformed both a linear classifier and a fine-tuned model by large margins, therefore we only reported the RF performances. 

In order to assess whether Deep Clustering could be relevant for learning spectral representations, we evaluated whether using spectral convolutions or dense layers would provide as strong prior on the input signal as convolutions for images. To this end, we evaluated the performances of a multi-layer perceptron trained on top of a randomly initialized 1) spectral CNN and 2) dense network.

Hyperparameters, including architecture details, learning rate, and weight decay, were selected through a random search on the validation set. Concerning the MAE that has a high number of hyperparameters, the masking ratio, the number of attention heads and the embedding dimension were selected through an ablation study presented in section \ref{sec:ablation}.

\subsubsection{Experimental results \label{sec:results}}

\begin{table*}[h]
\caption{Average overall accuracy and F1 score over the 8 standard ground truth splits \label{tab:acc_metrics}}
\begin{center}
\begin{tabular}{lcc}
\toprule
\textbf{Model} & \textbf{OA} & \textbf{F1 score} \\

KNN & 0.78 & 0.69 \\ 
AE + KNN & 0.82 & 0.73 \\
MAE + KNN & \textbf{0.84} & \textbf{0.76}
\end{tabular}
\hspace{2cm}
\begin{tabular}{lcc}
\toprule
\textbf{Model} & \textbf{OA} & \textbf{F1 score} \\

RF & 0.75 & 0.65 \\
AE + RF & 0.81 & 0.73 \\
MAE + RF & \textbf{0.85} & \textbf{0.77}

\end{tabular}
\end{center}
\end{table*}

\textbf{MAE} Results on Tab. \ref{tab:acc_metrics} show that the representations learned with an MAE combined with a KNN and a RF have led to a significant increase compared to a KNN (+7\% F1-score) and a RF (+12\% F1-score) applied on raw data, and to a standard autoencoder baseline, but by smaller margins. 

\textbf{Deep Clustering} A dense classifier trained on top of a random CNN and a random dense feature extractor achieved an average\footnote{F1-score averaged over the 8 ground truth splits.} F1-score of 0.038 and 0.024, respectively. For comparison, the average F1-score that a random classifier (with a uniform predictive distribution) would achieve on the Toulouse data set is 0.024. Therefore, the performances reached with a randomly initialized CNN and dense network are barely above chance and as good as chance, respectively. In conclusion, the use of spectral convolutions and dense layers to extract spectral representations do not provide useful priors on the spectral information, in contrast to convolutions for the spatial information of natural images. Thus, the pseudo-labels that we could derive from a clustering algorithm applied on the spectral features would not provide relevant semantic information. The iterative algorithm would not converge to useful spectral representations. This conclusion is confirmed by numerical experiments for which Deep Clustering combined with a KNN led to much worse accuracy than a KNN on the raw data.

\subsubsection{Ablation study \label{sec:ablation}}

We studied the impact of the masking ratio, the number of attention heads, and the dimension of the latent space on the MAE performance. Note that we speak indifferently of the latent space dimension and the total embedding dimension. We shall also precise that each element in the sequence is represented by an embedding whose actual dimension is the total embedding dimension divided by the number of attention heads. The experiments were conducted on different ground truth splits and for several random initializations of the MAE parameters. The mean and standard deviation of the validation loss and validation accuracy are reported for 20 epochs on Figs. \ref{fig:mask_ratio_ablation}, \ref{fig:n_heads} and \ref{fig:z_dim}.

\begin{minipage}{0.3\textwidth}
	\center
	\begin{minipage}{\textwidth}
		\center
		\includegraphics[width=\textwidth]{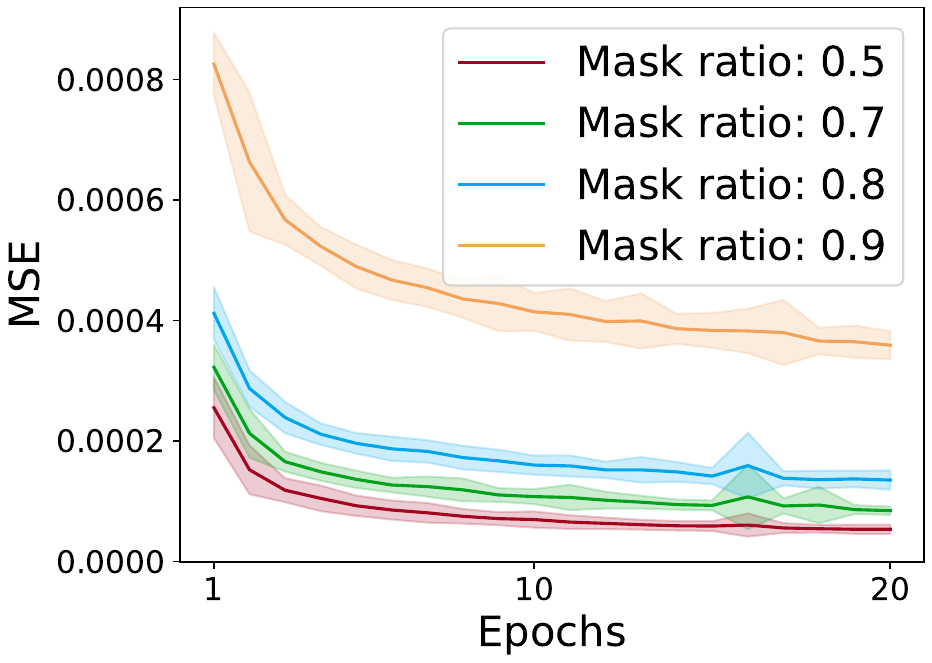}
	\end{minipage}
	
	\begin{minipage}{\textwidth}
		\center
		\includegraphics[width=\textwidth]{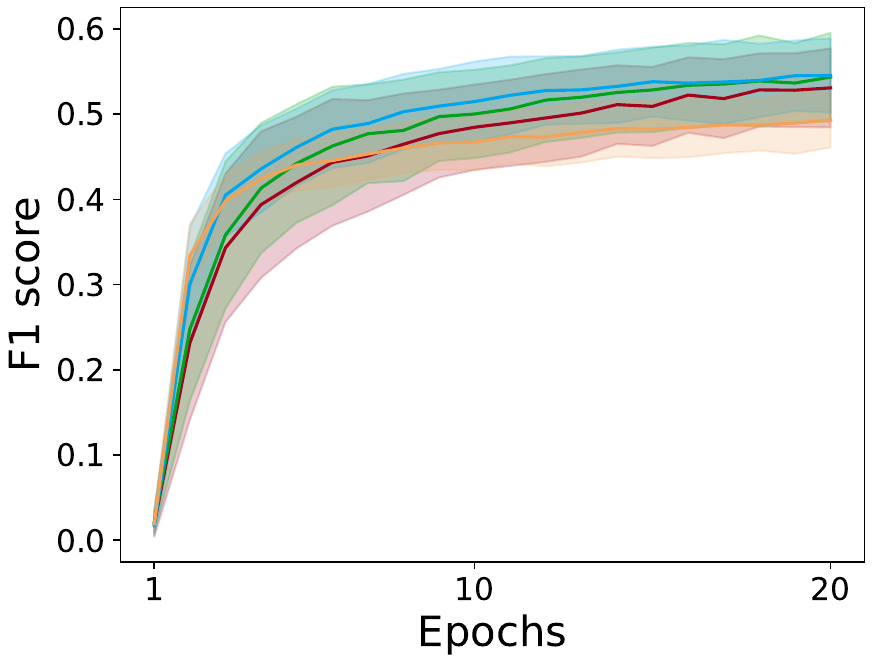}
	\end{minipage}
	\captionof{figure}{Influence of the masking ratio for a 32-dimensional latent space and 8 attention heads. Top: validation loss. Bottom: validation accuracy. \label{fig:mask_ratio_ablation}}
\end{minipage}
\hfill
\begin{minipage}{0.3\textwidth}
	\center
	\begin{minipage}{\textwidth}
		\center
		\includegraphics[width=\textwidth]{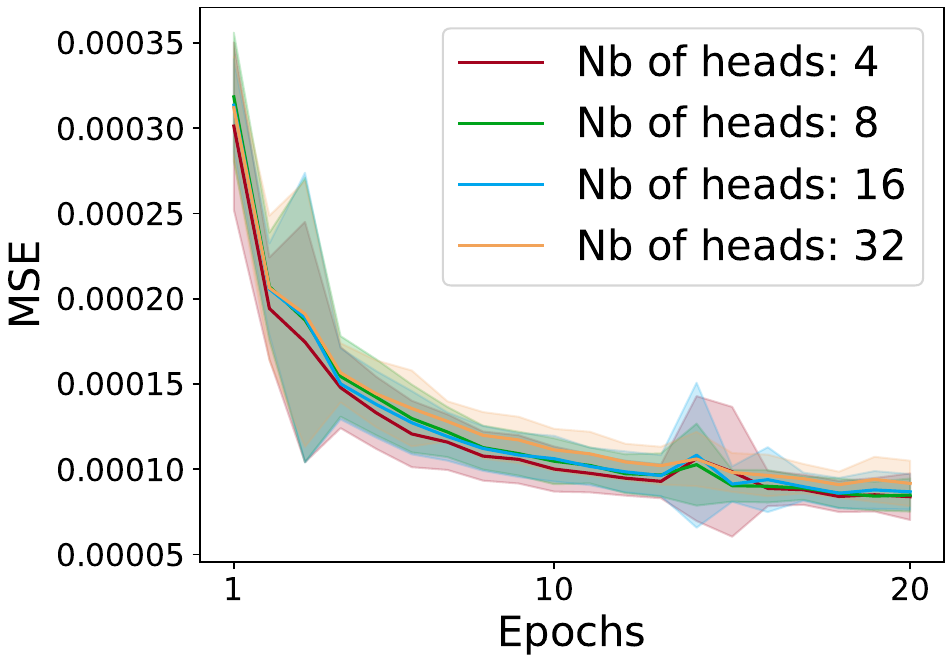}
	\end{minipage}
	\begin{minipage}{\textwidth}
		\center
		\includegraphics[width=\textwidth]{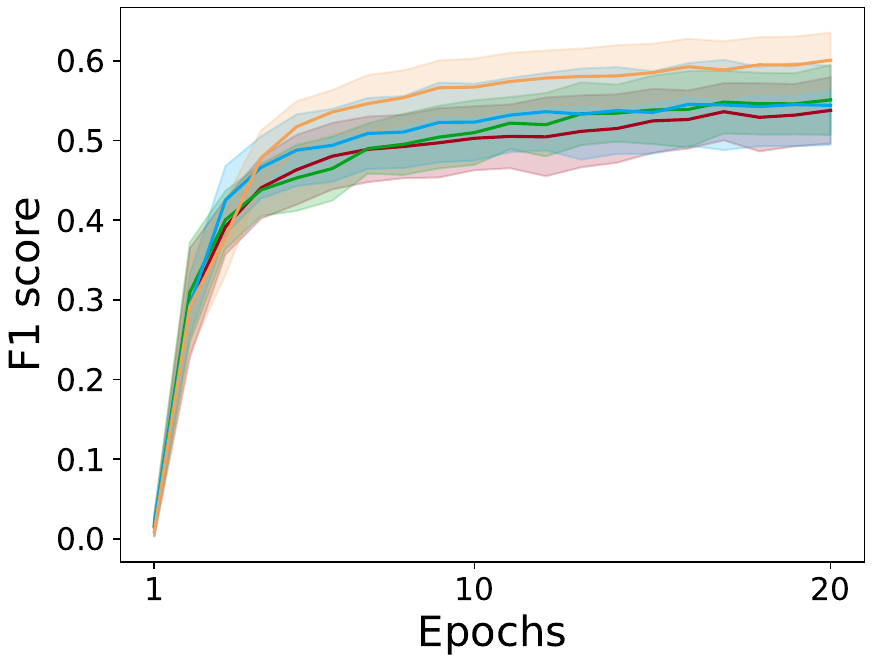}
	\end{minipage}
	\captionof{figure}{Influence of the number of attention heads for a 32-dimensional latent space and a 0.7 masking ratio. Top: validation loss. Bottom: validation accuracy. \label{fig:n_heads}}
\end{minipage}
\hfill
\begin{minipage}{0.3\textwidth}
	\center
	\begin{minipage}{\textwidth}
		\center
		\includegraphics[width=\textwidth]{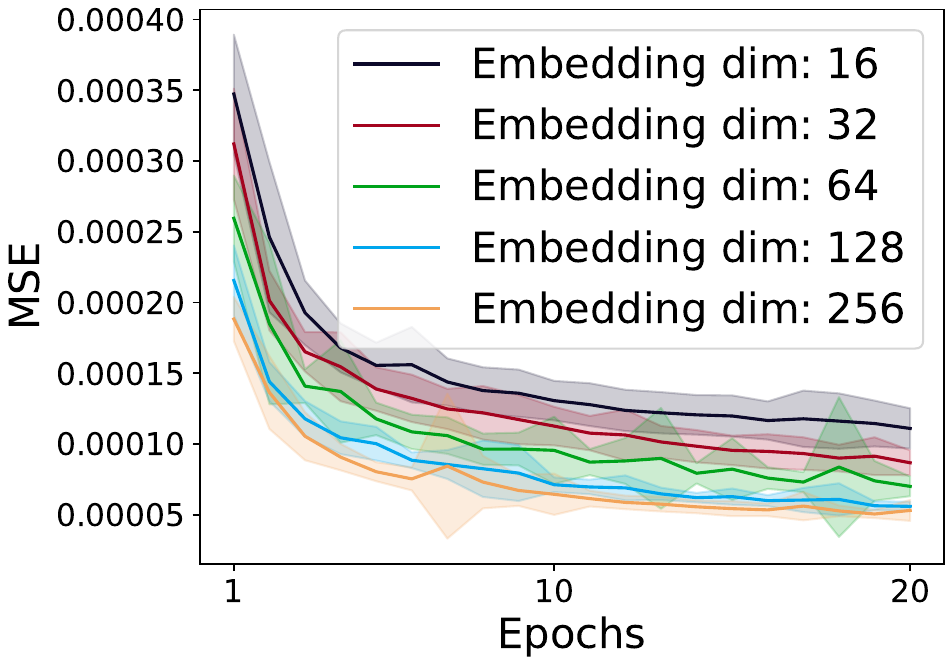}
	\end{minipage}
	\begin{minipage}{\textwidth}
		\center
		\includegraphics[width=\textwidth]{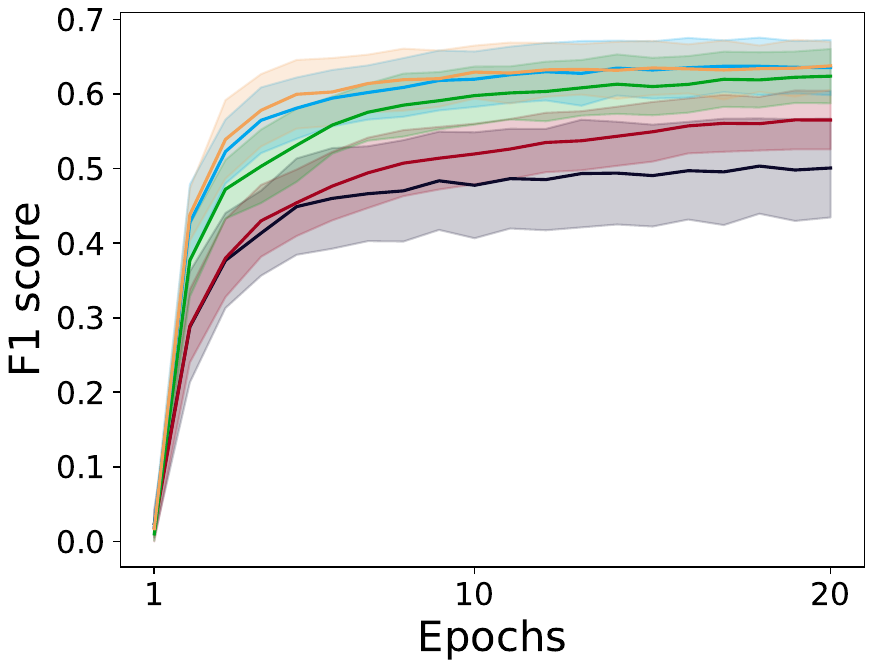}
	\end{minipage}
	\captionof{figure}{Influence of the latent space dimension for a 0.7 masking ratio and as many attention heads as latent space dimensions. Top: validation loss. Bottom: validation accuracy. \label{fig:z_dim}}
\end{minipage}

First, experiments on the masking ratio (Fig. \ref{fig:mask_ratio_ablation}) showed that a masking ratio around 0.7 and 0.8 leads to a trade-off between a trivial task and one that is too hard for the model to learn useful representations. Precisely, the lowest validation reconstruction error is obtained with a 0.5 masking ratio but the highest validation accuracies are obtained with 0.7 and 0.8 masking ratios. 
 
Second, experiments on the number of attention heads (Fig. \ref{fig:n_heads}) showed that best performances were reached when the number of attention heads equaled the embedding dimension, \textit{i.e} when each element in the sequence had a 1-dimensional (scalar) representation. While the validation reconstruction errors were bearly unchanged, a significant increase was reached with 32 attention heads. This result is consistent with our prior belief that spectral data, in contrast to words, contain very low-level information, and therefore do not require high-dimensional embeddings.

Third, experiments on the latent space dimension (Fig. \ref{fig:z_dim}) showed that best validation loss and validation accuracies were reached for 128 and 256 dimensions, despite the increase of parameters in the encoder.

\section{Conclusions and perspectives \label{sec:conclusion}}

We have introduced the Toulouse Hyperspectral Data Set, a large benchmark data set designed to assess semi/self-supervised representation learning and pixel-wise classification techniques. Quantitative and qualitative comparisons have shown that several properties of the Toulouse data set better reflect the complexity of the land cover in large urban areas, compared to currently public data sets.
In order to facilitate fair and reproducible experiments, we released a Python library to easily load PyTorch\footnote{https://pytorch.org/docs/stable/index.html} data loaders and we hope that the standard train / test ground truth splits will foster the evaluation and comparison of new model architectures and learning algorithms.

The numerical experiments showed that the masked autoencoding task is very appropriate for learning useful spectral representations. The baseline established in this paper for pixel-wise classification based on a MAE \cite{he2022masked} and Random Forests \cite{breiman2001random} reached a 85\% overall accuracy and 77\% F1 score. Besides, the ablation study about the MAE hyperparameters showed that a higher number of attention heads (and a lower embedding dimension for each element in the sequence) than for vision transformers is beneficial to extract spectral features. We argue that this result can be explained by the fact that reflectance spectra contain low-level information, compared to RGB images. Experiments also demonstrated that the use of spectral convolutions do not provide as strong prior on reflectance spectra as convolutions do for RGB images, which explains why clustering-based techniques are inadequate for spectral representation learning.

We focused in this paper on the spectral information because the reflectance, that is intrinsic to the chemical composition of matter, is by nature highly discriminating of the land cover. In some cases, however, we believe that large-scale contextual information could prevent some confusions (for instance, pixels predicted as \textit{wheat} in a green urban area or pixels predicted as \textit{orange tile} in a cultivated field). Precisely, we argue that hyperspectral patches of at least $64 \times 64$ pixels are necessary with a $\approx$ 1 m GSD (while common semantic segmentation models applied on RGB satellite images rather use $256 \times 256$ pixel patches with a $\approx$ 50 cm GSD). Processing such big hyperspectral patches with macine learning models though raises memory issues due to the large spectral dimension, which hinders the direct application of state-of-the-art vision models, especially large models such as the recent foundation model SAM \cite{kirillov2023segment}. Nevertheless, combining deep vision models to extract contextual information (\textit{i.e.} abstract land use information) with shallow models to extract spectral information (\textit{i.e.} low-level land cover information) is a promising research direction. Besides, we believe that the hierarchical nomenclature, the land use annotations in addition to the land cover annotations and the long-tailed class distribution open the path towards important research areas, respectively (hierarchical) multi-label classification \cite{vens2008decision} and long-tailed learning \cite{zhang2023deep}, that have been little discussed for pixel-wise hyperspecral image classification. Lastly, the Toulouse Hyperspectral Data Set is also particularly suited to evaluate Active Learning algorithms thanks to the provided \textit{labeled pools} from which pixels to label can be sampled.

\section*{Acknowledgement}

We thank Philippe Déliot for providing the hyperspectral images in ground-level reflectance and geometrically rectified.

\clearpage
\bibliographystyle{unsrt}  
\bibliography{references}  

\clearpage
\begin{appendices}

\Large{\textbf{Appendices}}

\vspace{1cm}

\begin{minipage}{\textwidth}
	\center
	\begin{minipage}{0.3\textwidth}
		\center
		\includegraphics[width=\textwidth]{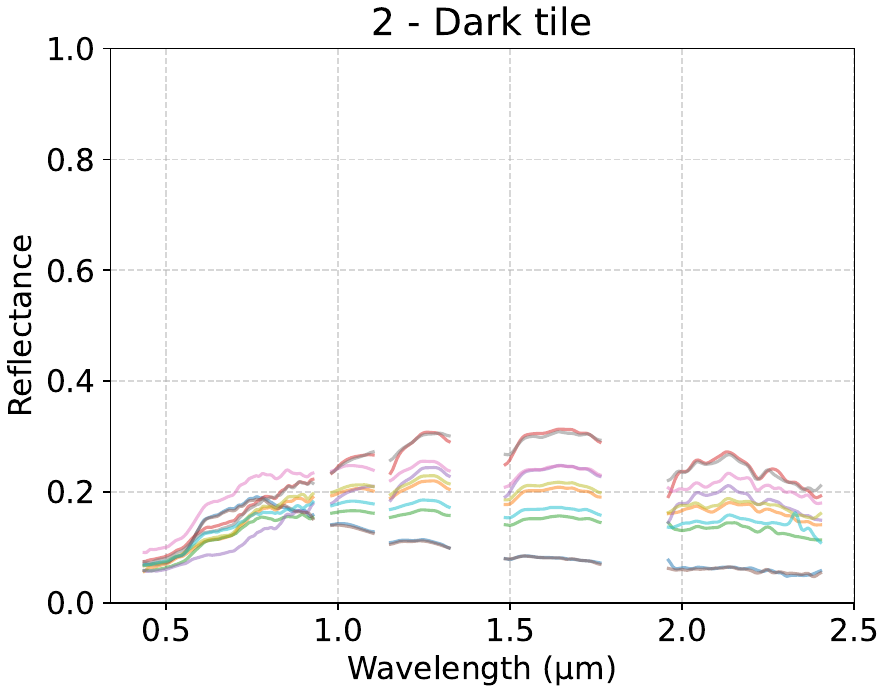}
	\end{minipage}
	\hfill
	\begin{minipage}{0.3\textwidth}
		\center
		\includegraphics[width=\textwidth]{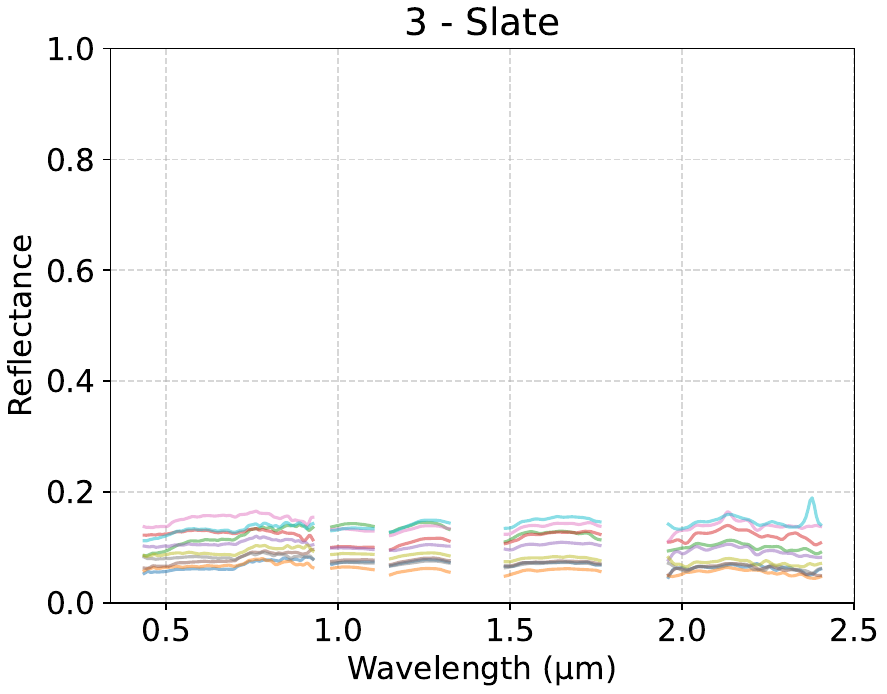}
	\end{minipage}
	\hfill
	\begin{minipage}{0.3\textwidth}
		\center
		\includegraphics[width=\textwidth]{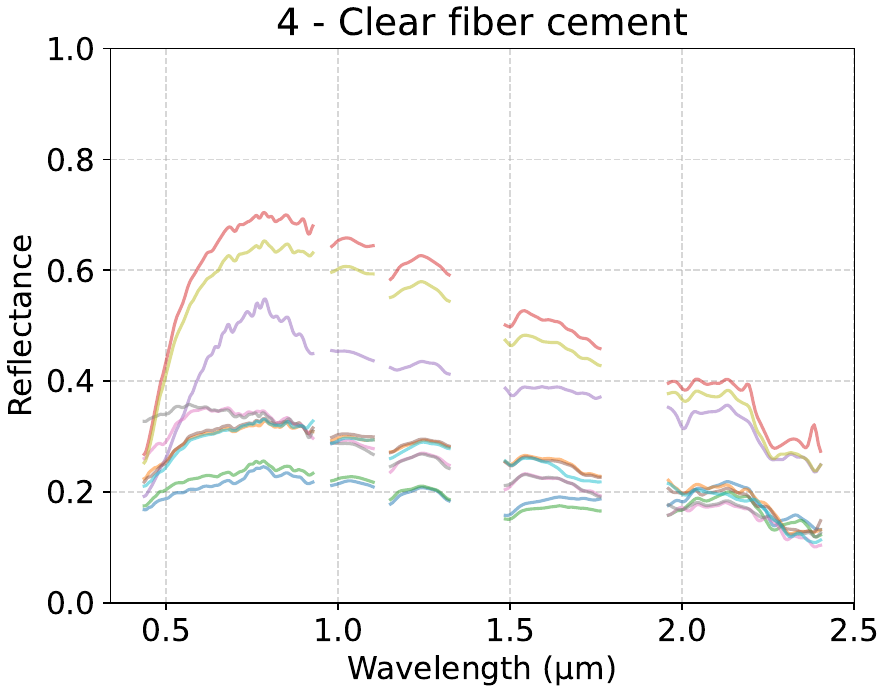}
	\end{minipage}
	
	\begin{minipage}{0.3\textwidth}
		\center
		\includegraphics[width=\textwidth]{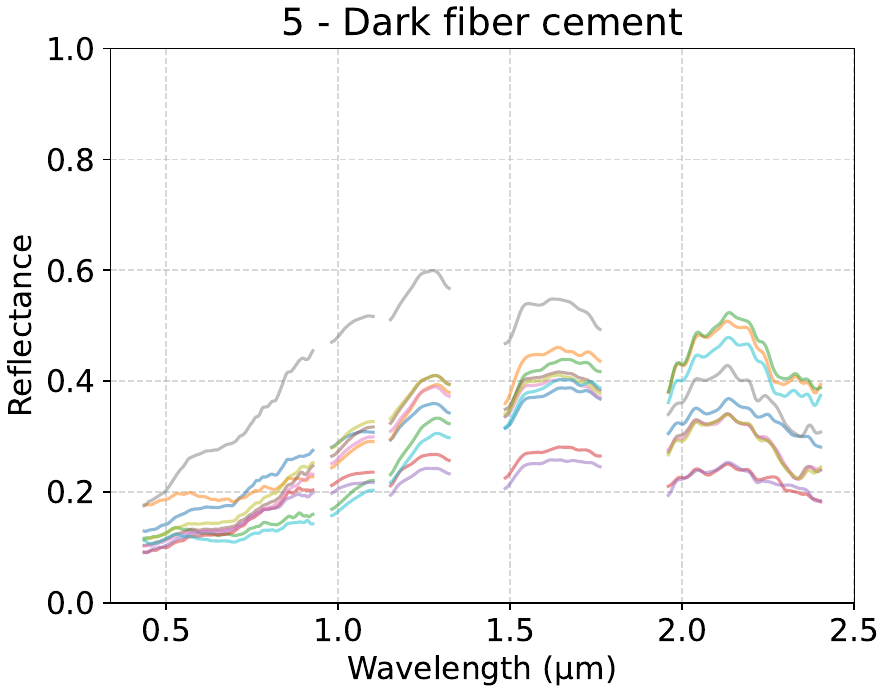}
	\end{minipage}
	\hfill
	\begin{minipage}{0.3\textwidth}
		\center
		\includegraphics[width=\textwidth]{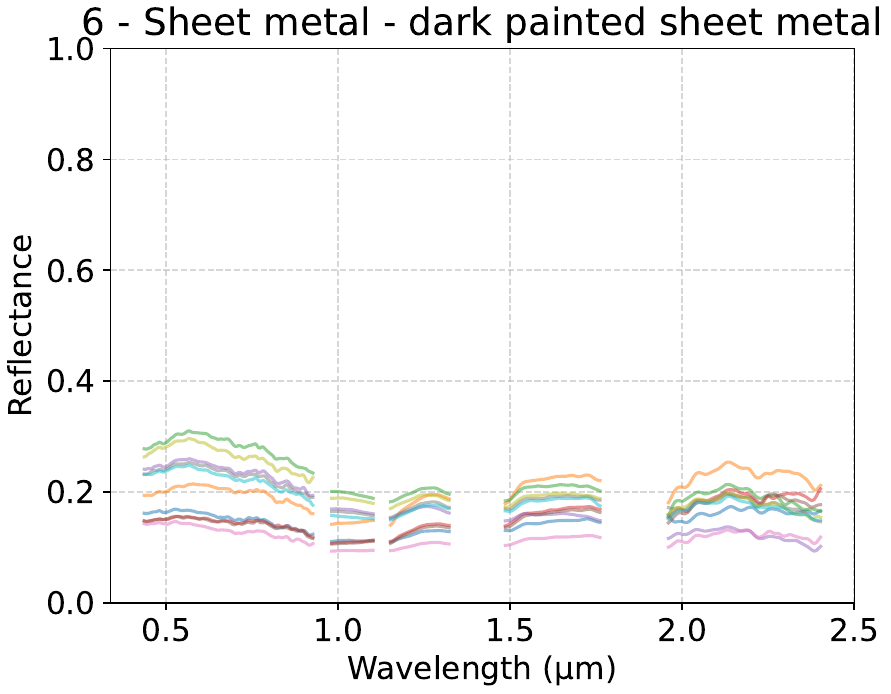}
	\end{minipage}
	\hfill
	\begin{minipage}{0.3\textwidth}
		\center
		\includegraphics[width=\textwidth]{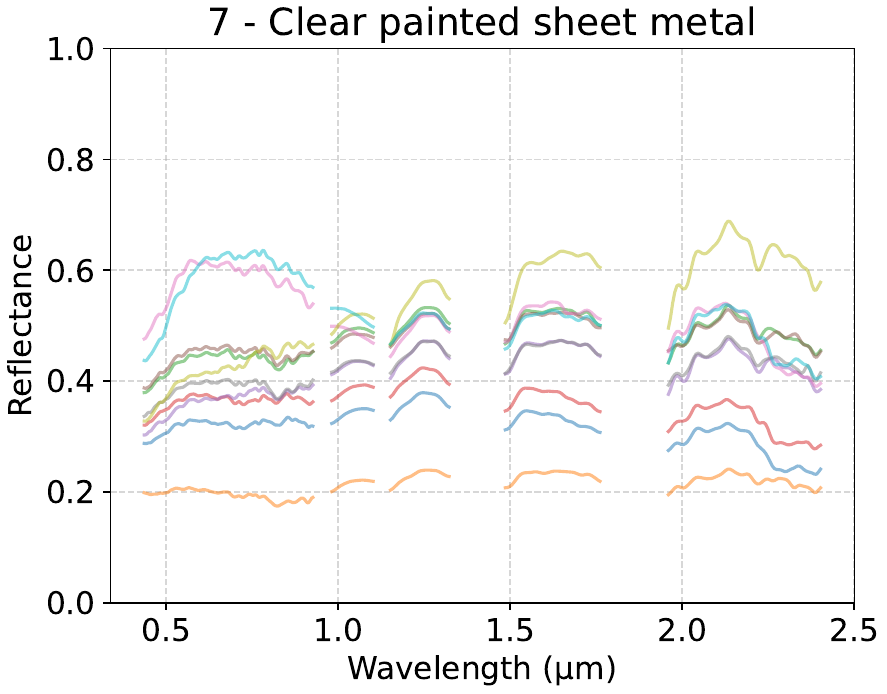}
	\end{minipage}
	
	\begin{minipage}{0.3\textwidth}
		\center
		\includegraphics[width=\textwidth]{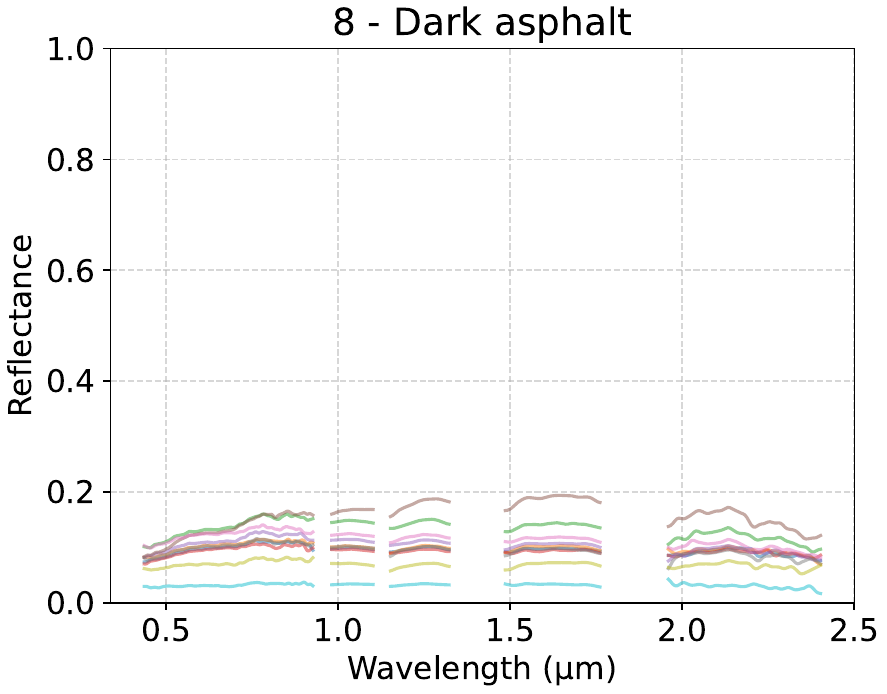}
	\end{minipage}
	\hfill
	\begin{minipage}{0.3\textwidth}
		\center
		\includegraphics[width=\textwidth]{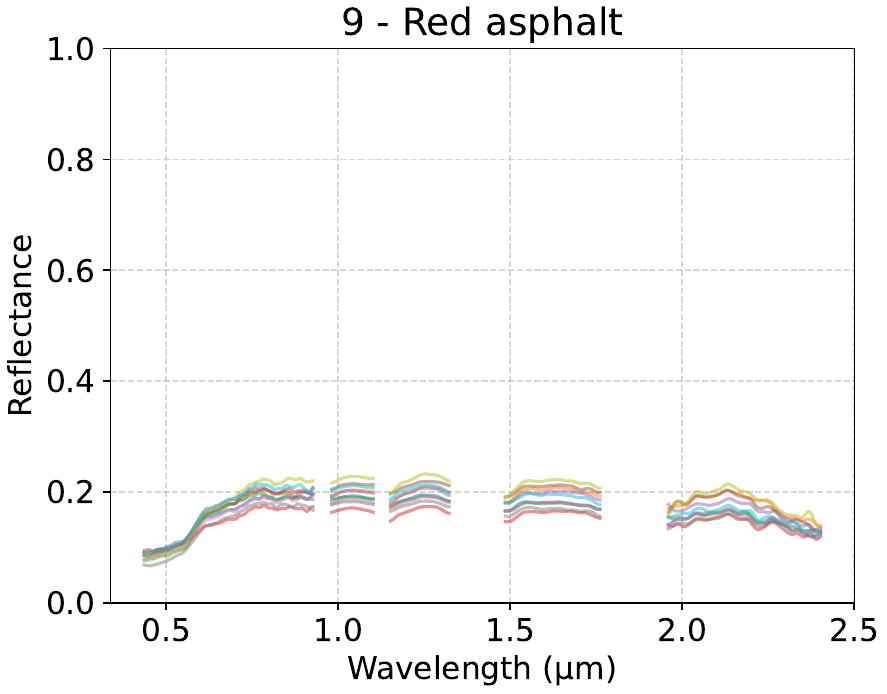}
	\end{minipage}
	\hfill
	\begin{minipage}{0.3\textwidth}
		\center
		\includegraphics[width=\textwidth]{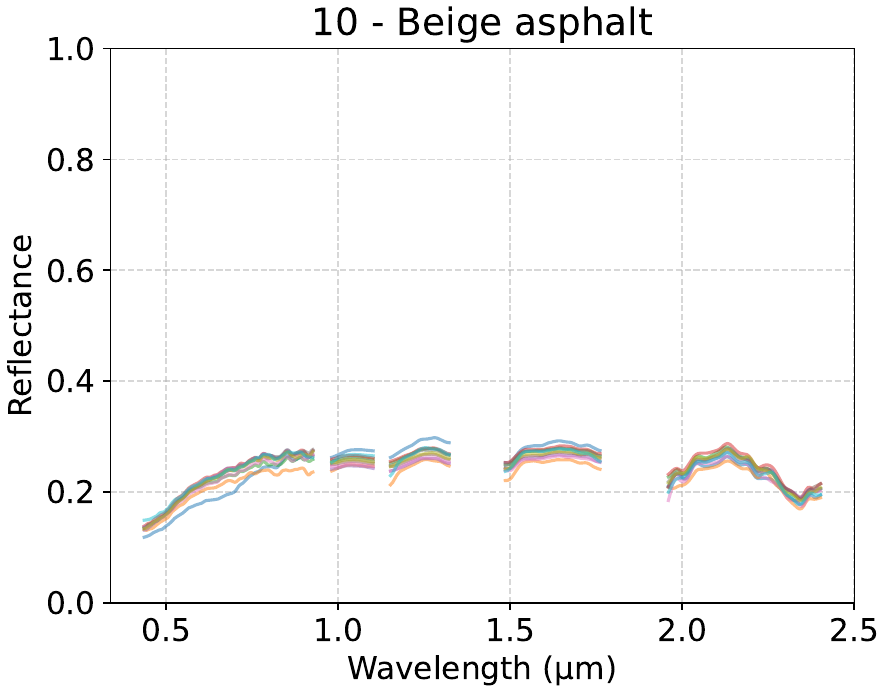}
	\end{minipage}

	\begin{minipage}{0.3\textwidth}
		\center
		\includegraphics[width=\textwidth]{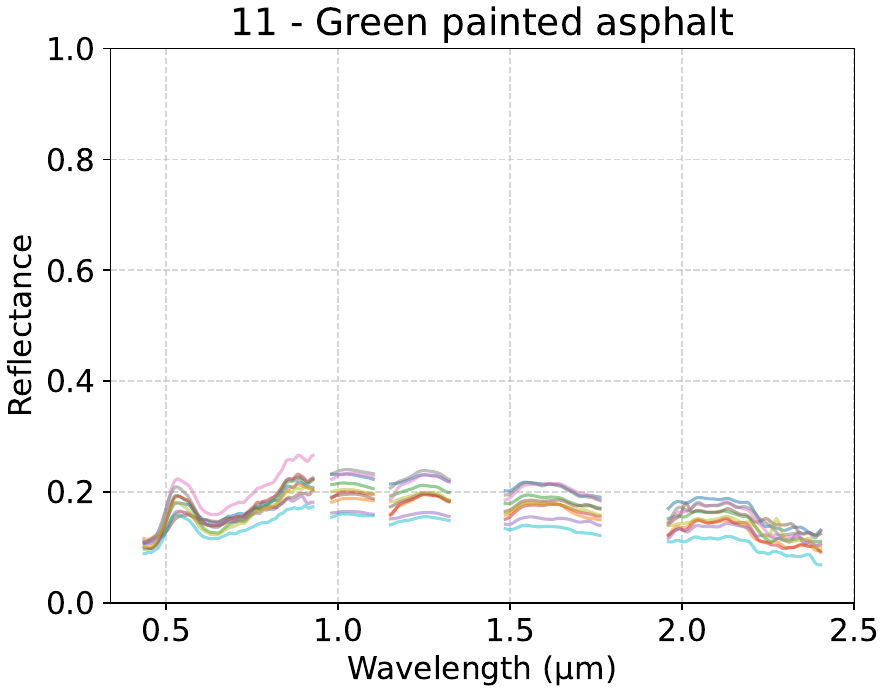}
	\end{minipage}
	\hfill
	\begin{minipage}{0.3\textwidth}
		\center
		\includegraphics[width=\textwidth]{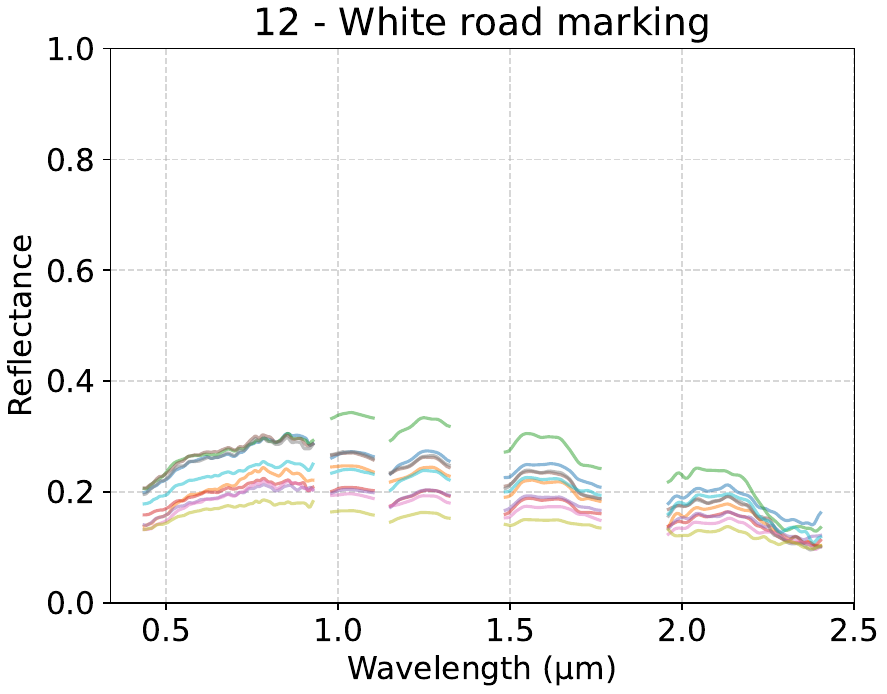}
	\end{minipage}
	\hfill
	\begin{minipage}{0.3\textwidth}
		\center
		\includegraphics[width=\textwidth]{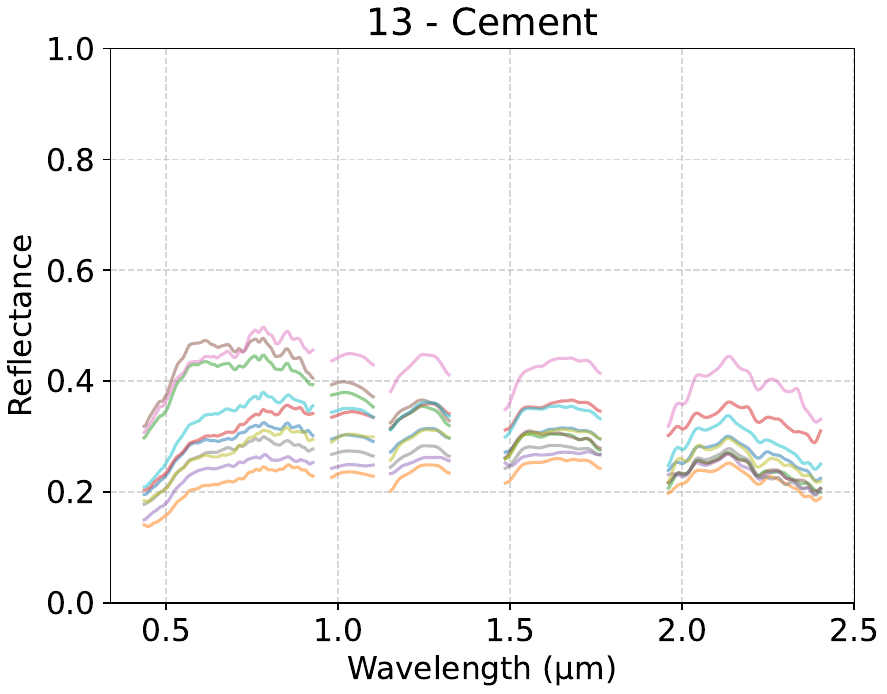}
	\end{minipage}
	
	\begin{minipage}{0.3\textwidth}
		\center
		\includegraphics[width=\textwidth]{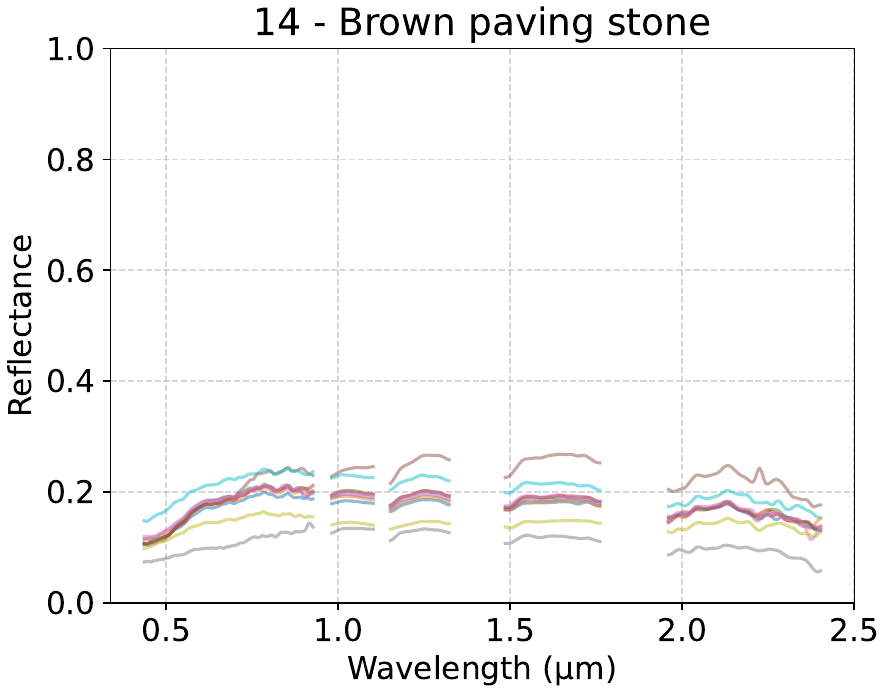}
	\end{minipage}
	\hfill
	\begin{minipage}{0.3\textwidth}
		\center
		\includegraphics[width=\textwidth]{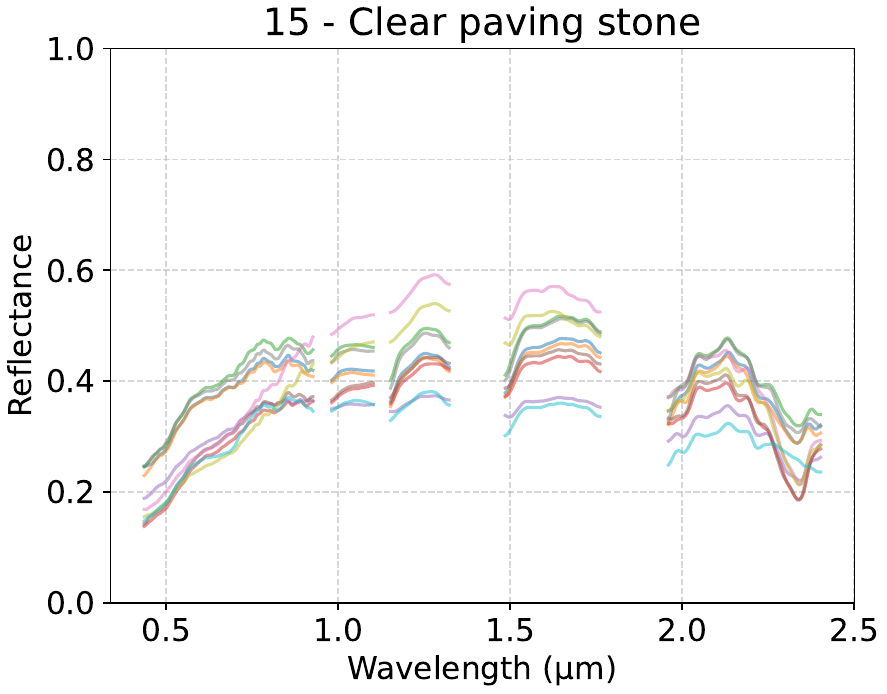}
	\end{minipage}
	\hfill
	\begin{minipage}{0.3\textwidth}
		\center
		\includegraphics[width=\textwidth]{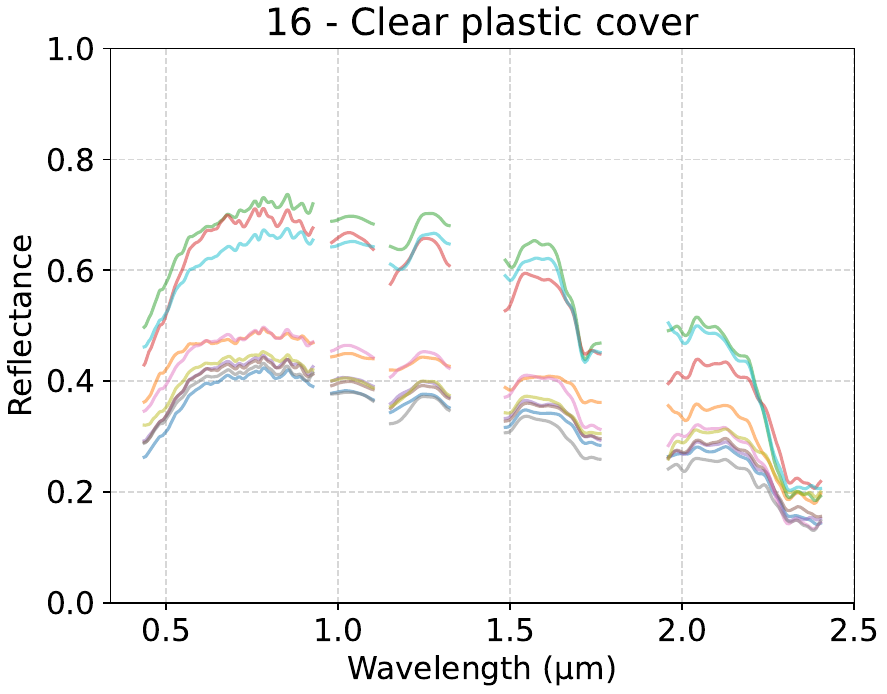}
	\end{minipage}
\end{minipage}

\begin{minipage}{\textwidth}
	\center
	\begin{minipage}{0.3\textwidth}
		\center
		\includegraphics[width=\textwidth]{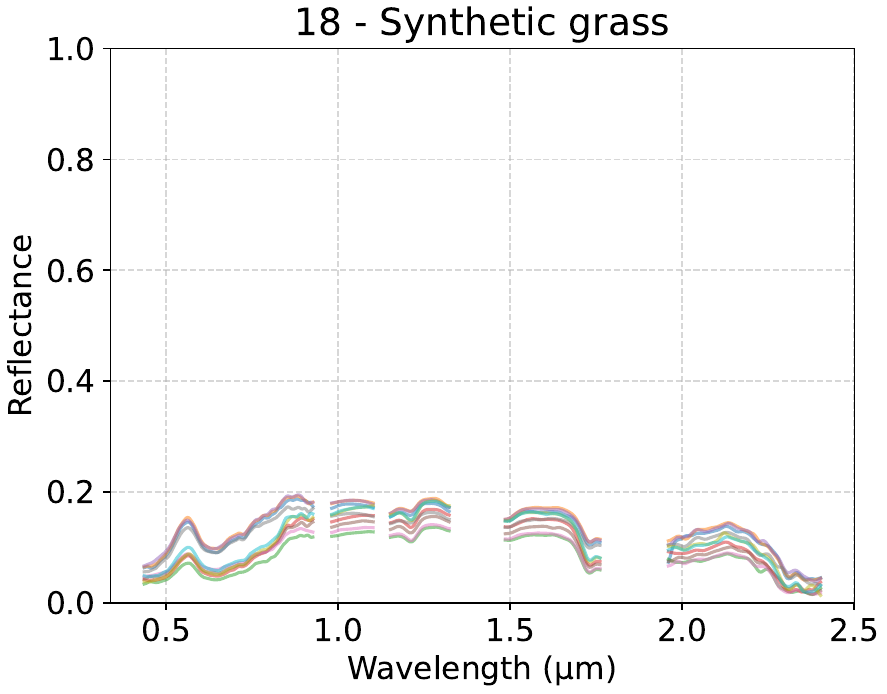}
	\end{minipage}
	\hfill
	\begin{minipage}{0.3\textwidth}
		\center
		\includegraphics[width=\textwidth]{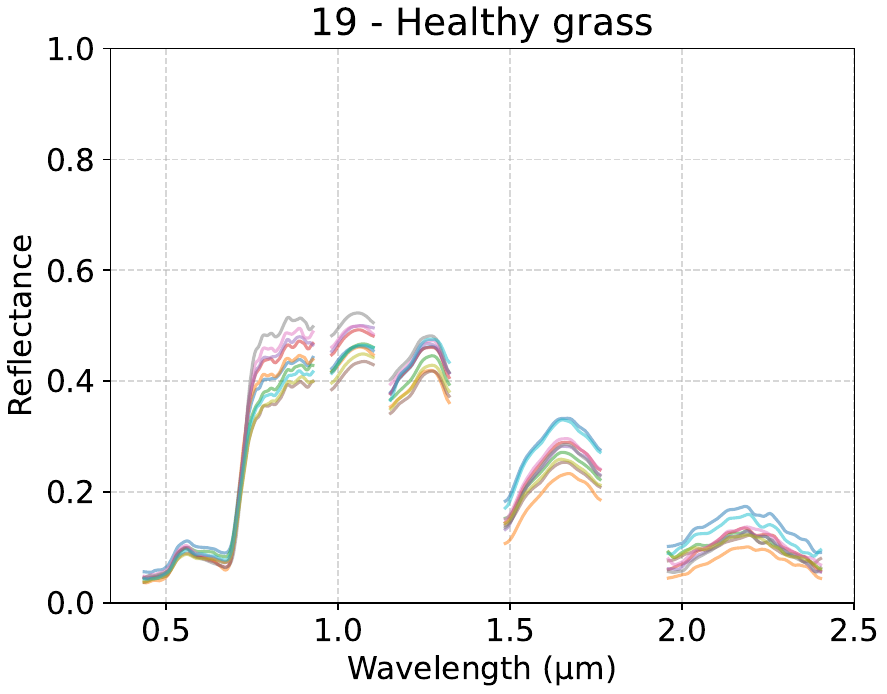}
	\end{minipage}
	\hfill
	\begin{minipage}{0.3\textwidth}
		\center
		\includegraphics[width=\textwidth]{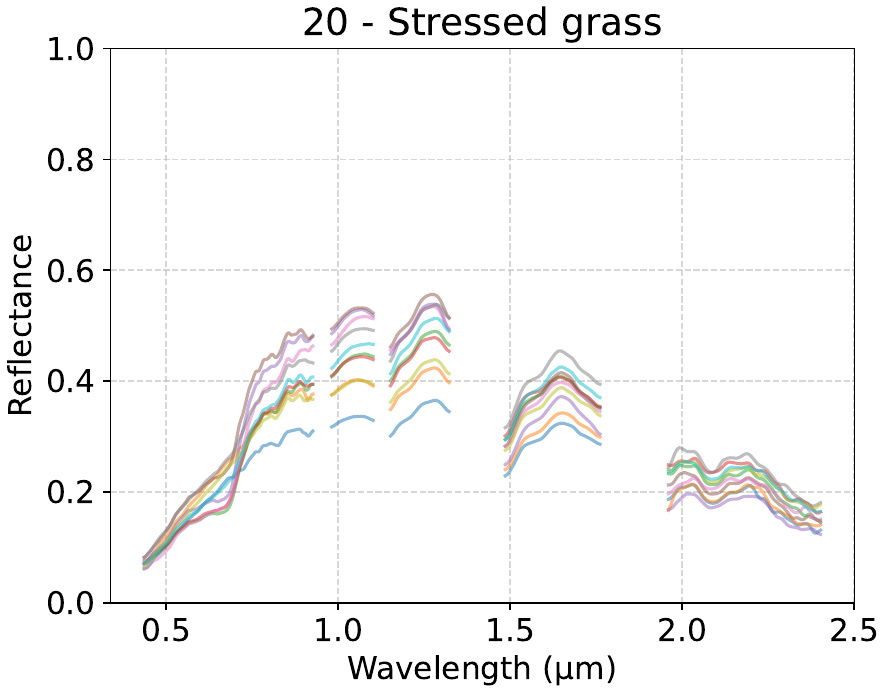}
	\end{minipage}
	
	\begin{minipage}{0.3\textwidth}
		\center
		\includegraphics[width=\textwidth]{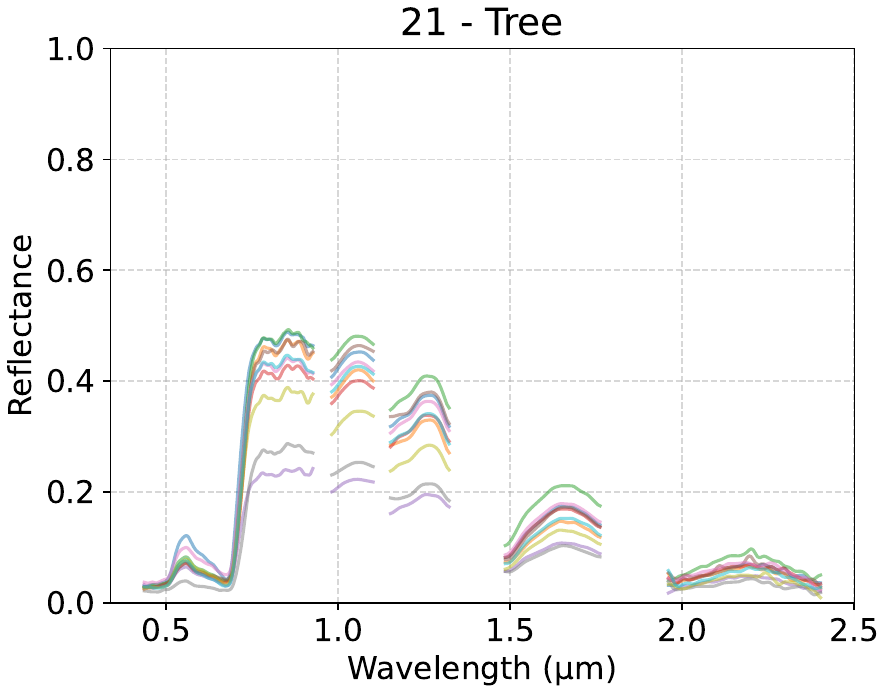}
	\end{minipage}
	\hfill
	\begin{minipage}{0.3\textwidth}
		\center
		\includegraphics[width=\textwidth]{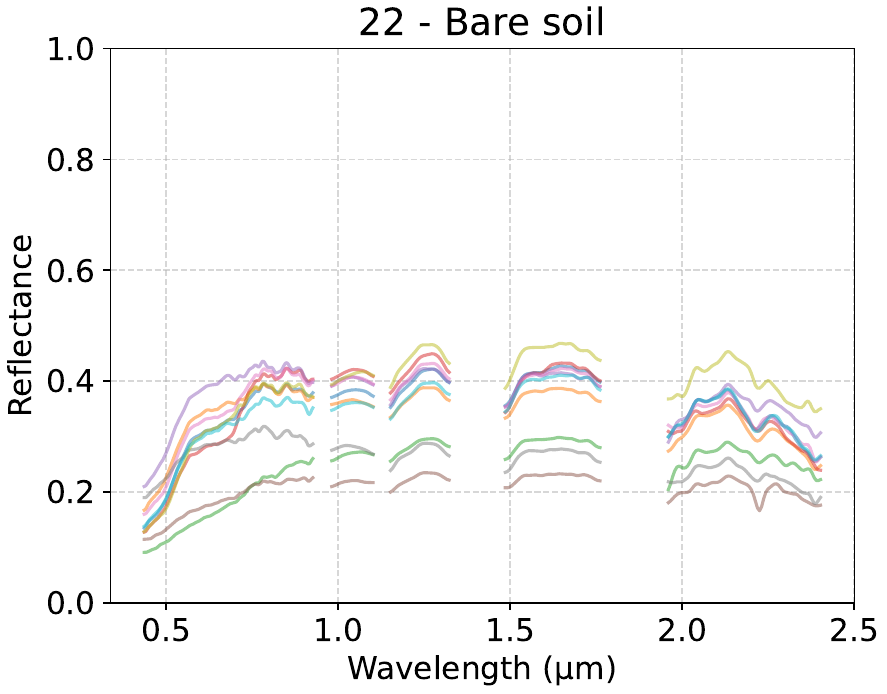}
	\end{minipage}
	\hfill
	\begin{minipage}{0.3\textwidth}
		\center
		\includegraphics[width=\textwidth]{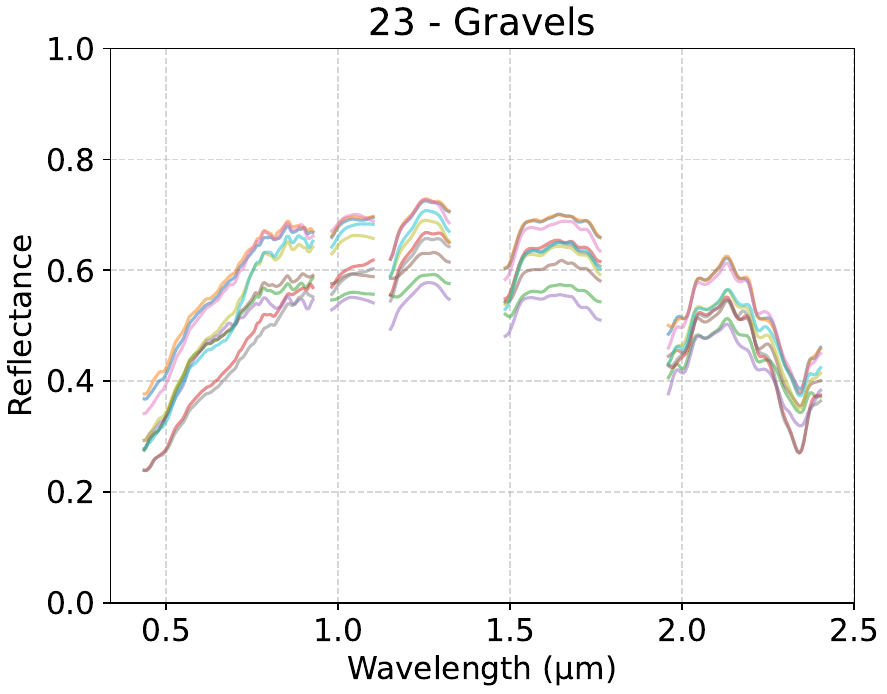}
	\end{minipage}

	\begin{minipage}{0.3\textwidth}
		\center
		\includegraphics[width=\textwidth]{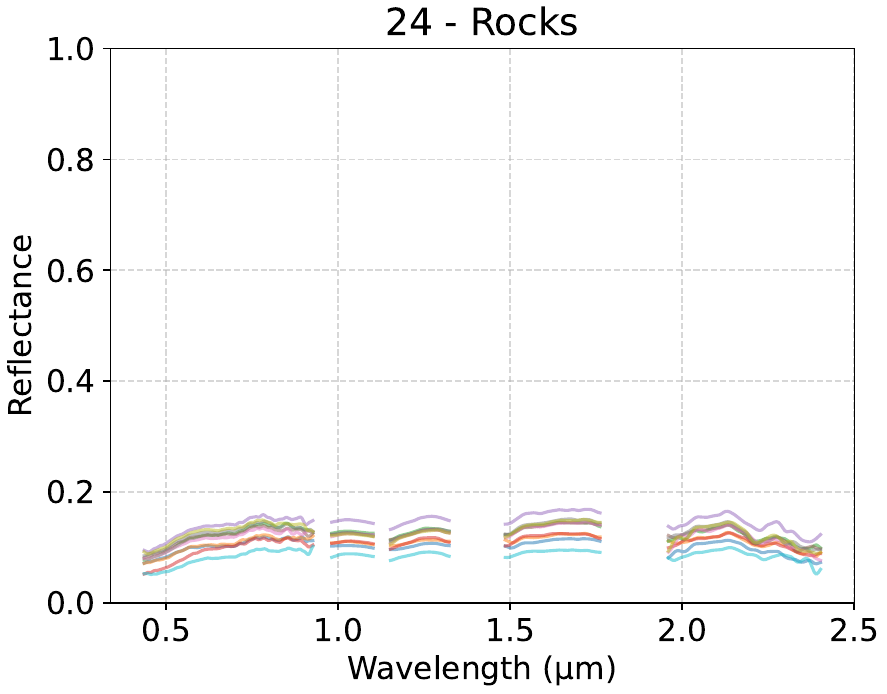}
	\end{minipage}
	\hfill
	\begin{minipage}{0.3\textwidth}
		\center
		\includegraphics[width=\textwidth]{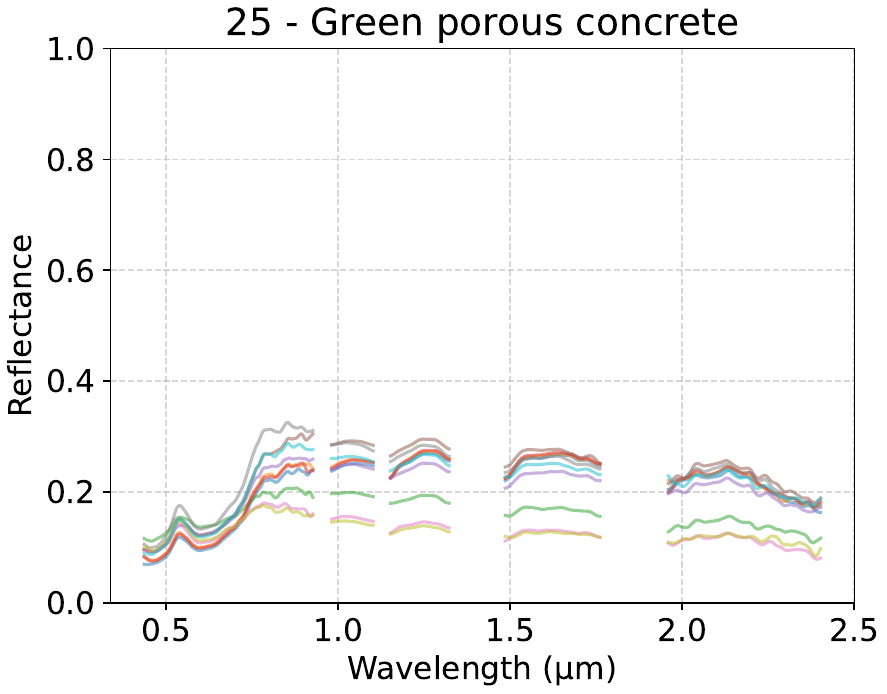}
	\end{minipage}
	\hfill
	\begin{minipage}{0.3\textwidth}
		\center
		\includegraphics[width=\textwidth]{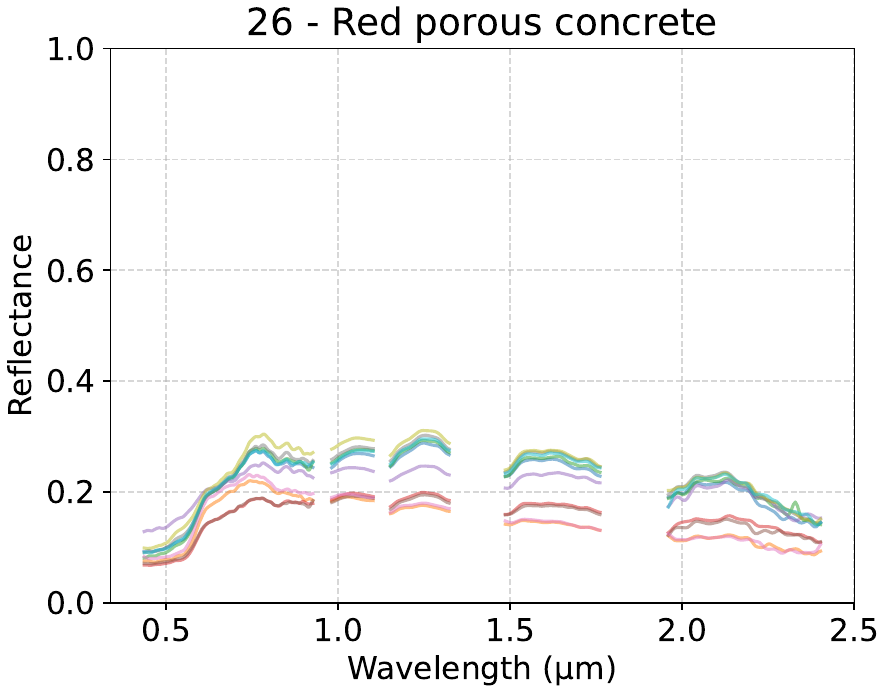}
	\end{minipage}
	
	\begin{minipage}{0.3\textwidth}
		\center
		\includegraphics[width=\textwidth]{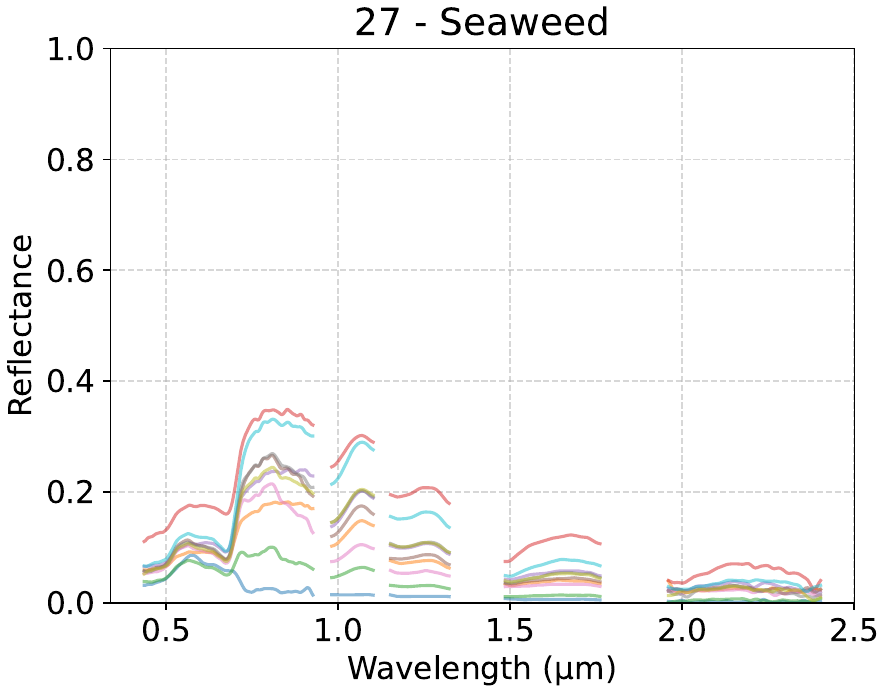}
	\end{minipage}
	\hfill
	\begin{minipage}{0.3\textwidth}
		\center
		\includegraphics[width=\textwidth]{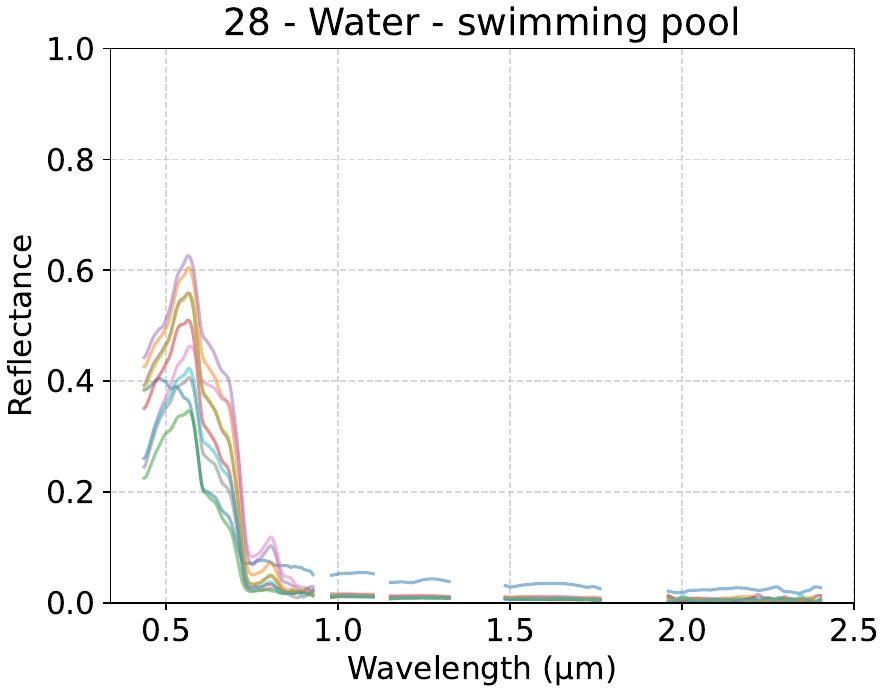}
	\end{minipage}
	\hfill
	\begin{minipage}{0.3\textwidth}
		\center
		\includegraphics[width=\textwidth]{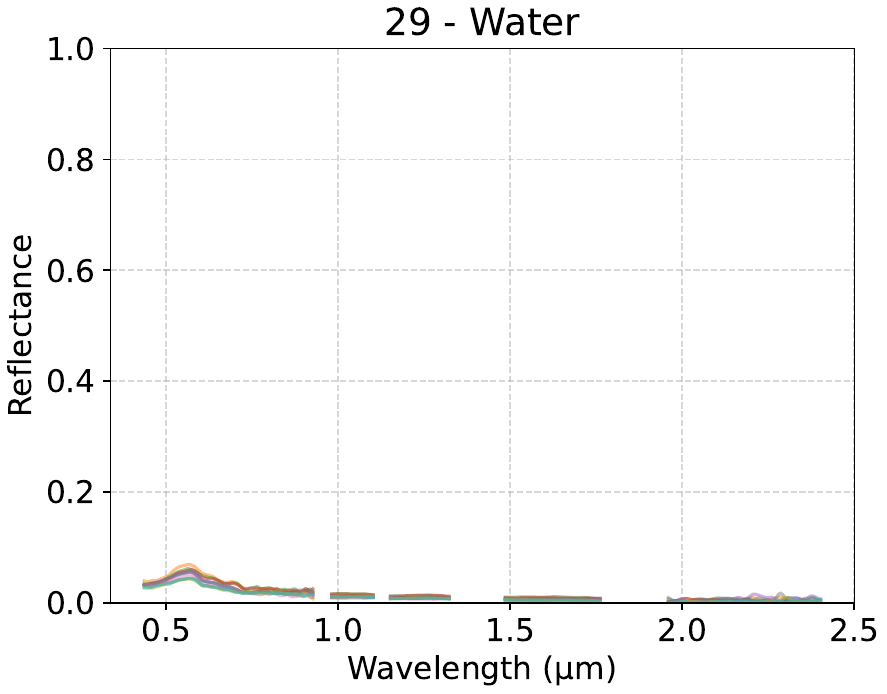}
	\end{minipage}
	
	\begin{minipage}{0.3\textwidth}
		\center
		\includegraphics[width=\textwidth]{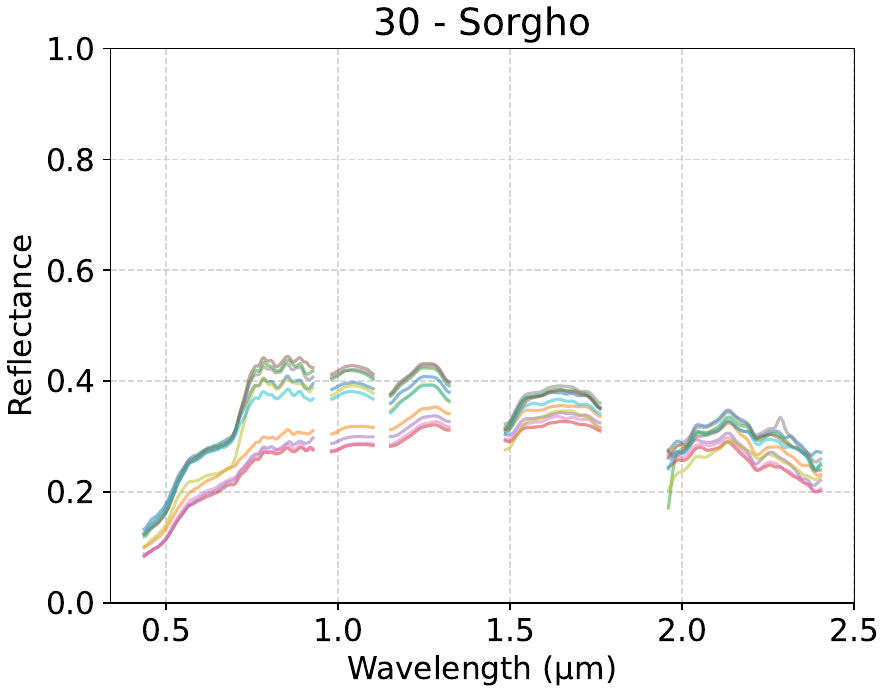}
	\end{minipage}
	\hfill
	\begin{minipage}{0.3\textwidth}
		\center
		\includegraphics[width=\textwidth]{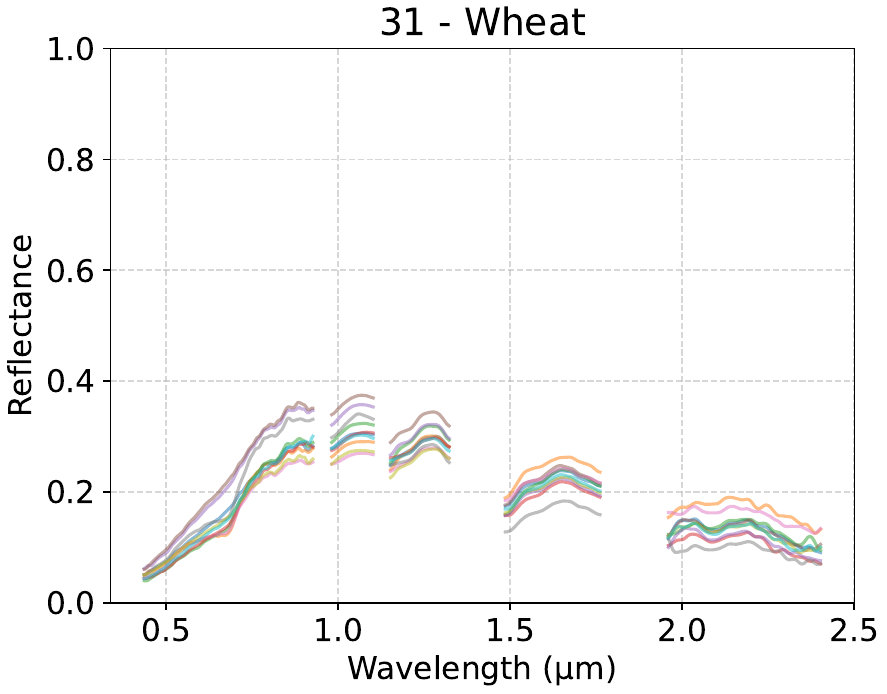}
	\end{minipage}
	\hfill
	\begin{minipage}{0.3\textwidth}
		\center
		\includegraphics[width=\textwidth]{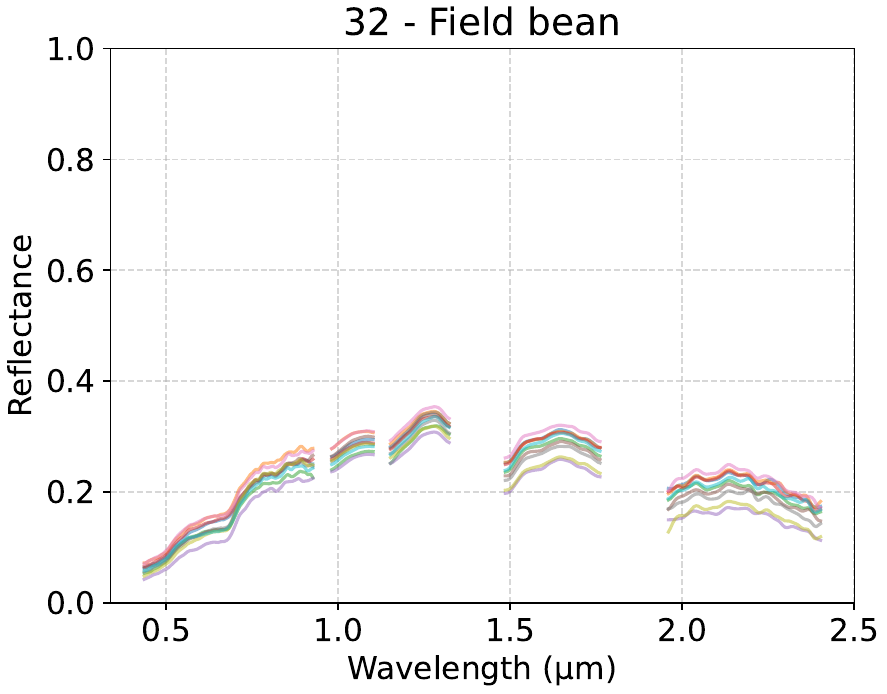}
	\end{minipage}
	\captionof{figure}{Additional random spectra of the Toulouse Hyperspectral Data Set \label{fig:spectra}}
\end{minipage}

\clearpage 


\normalsize{In addition to the land cover, we define a land use nomenclature which gathers more abstract semantic classes, listed in Tab. \ref{tab:land_use_classes}. Besides, we provide the direct and diffuse irradiance at ground level, as well as the solar zenith angle which is of 22.12°.}

\begin{table}[h]
	\caption{Land use nomenclature of the Toulouse Hyperspectral Data Set \label{tab:land_use_classes}}
	\begin{center}
	\begin{tabular}{llll}
		\toprule
		\#1 & Roads & \#7 & Lakes / rivers / harbors\\
		\#2 & Railways & \#8 & Swimming pools\\
		\#3 & Roofs & \#9 & Forests\\
		\#4 & Parking lots & \#10 & Cultivated fields\\
		\#5 & Building sites & \#11 & Boats \\
		\#6 & Sport facilities & \#12 & Open areas \\
	\end{tabular}
	\end{center}
\end{table}

\begin{figure}[h]
	\center
	\includegraphics[width=0.4\textwidth]{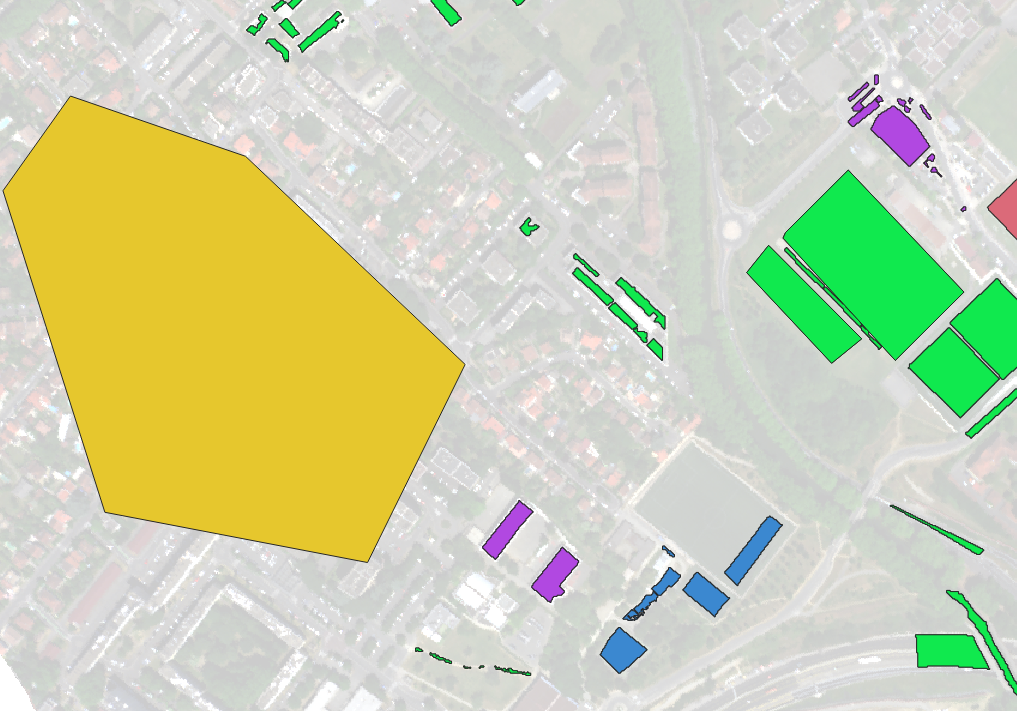}
	\caption{Example of annotations of the hyperspectral image. Polygons in red, green, blue, purple and yellow belong to the train labeled set, the labeled pool, the validation set, the test set and the unlabeled pool, respectively. \label{fig:polygons}}
\end{figure} 

\begin{figure}[h]
	\center	
	\begin{subfigure}{0.45\textwidth}
		\center
		\includegraphics[width=0.8\textwidth]{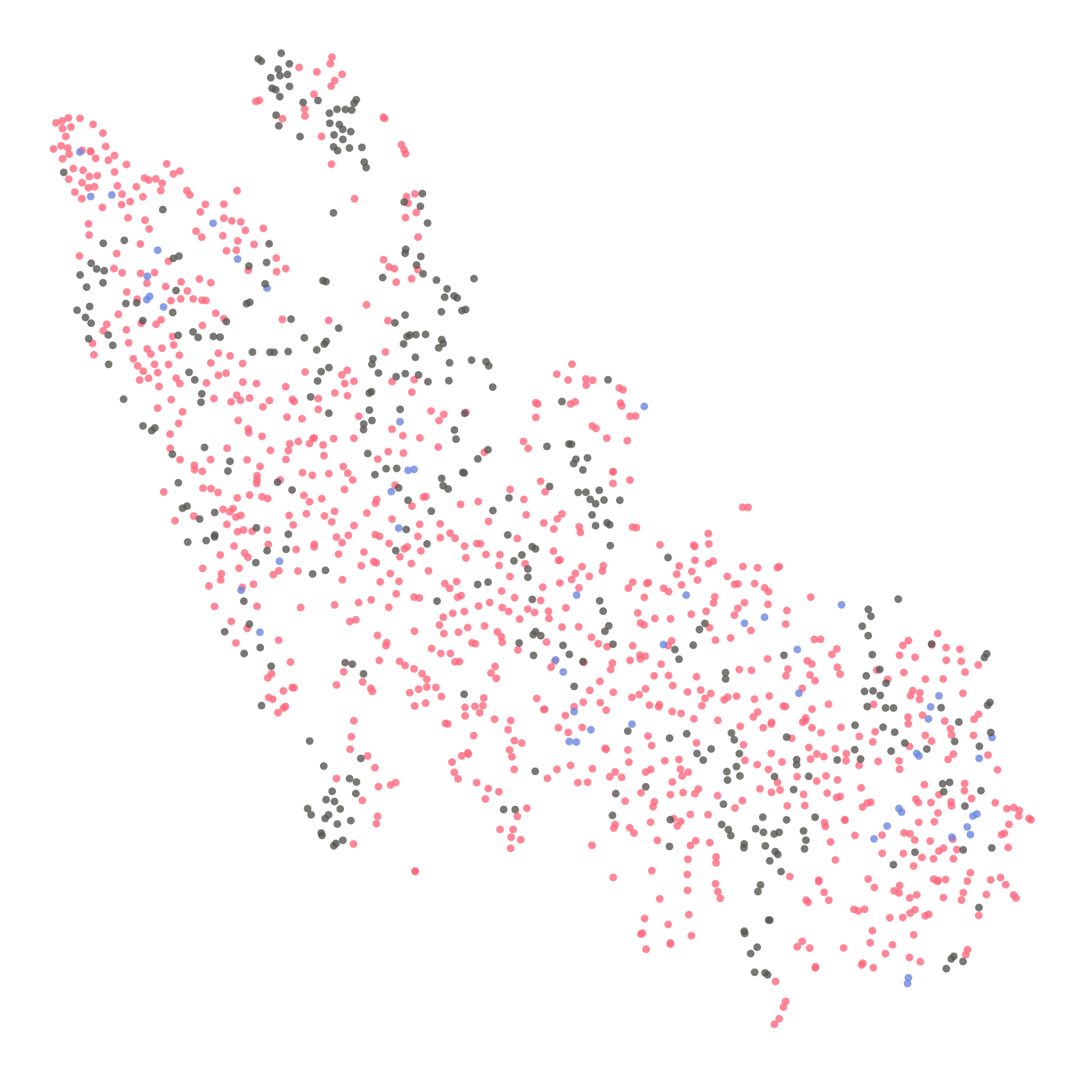}
	\caption{Spatial features}
	\end{subfigure}
	\hfill
	\begin{subfigure}{0.45\textwidth}
		\center
		\includegraphics[width=0.8\textwidth]{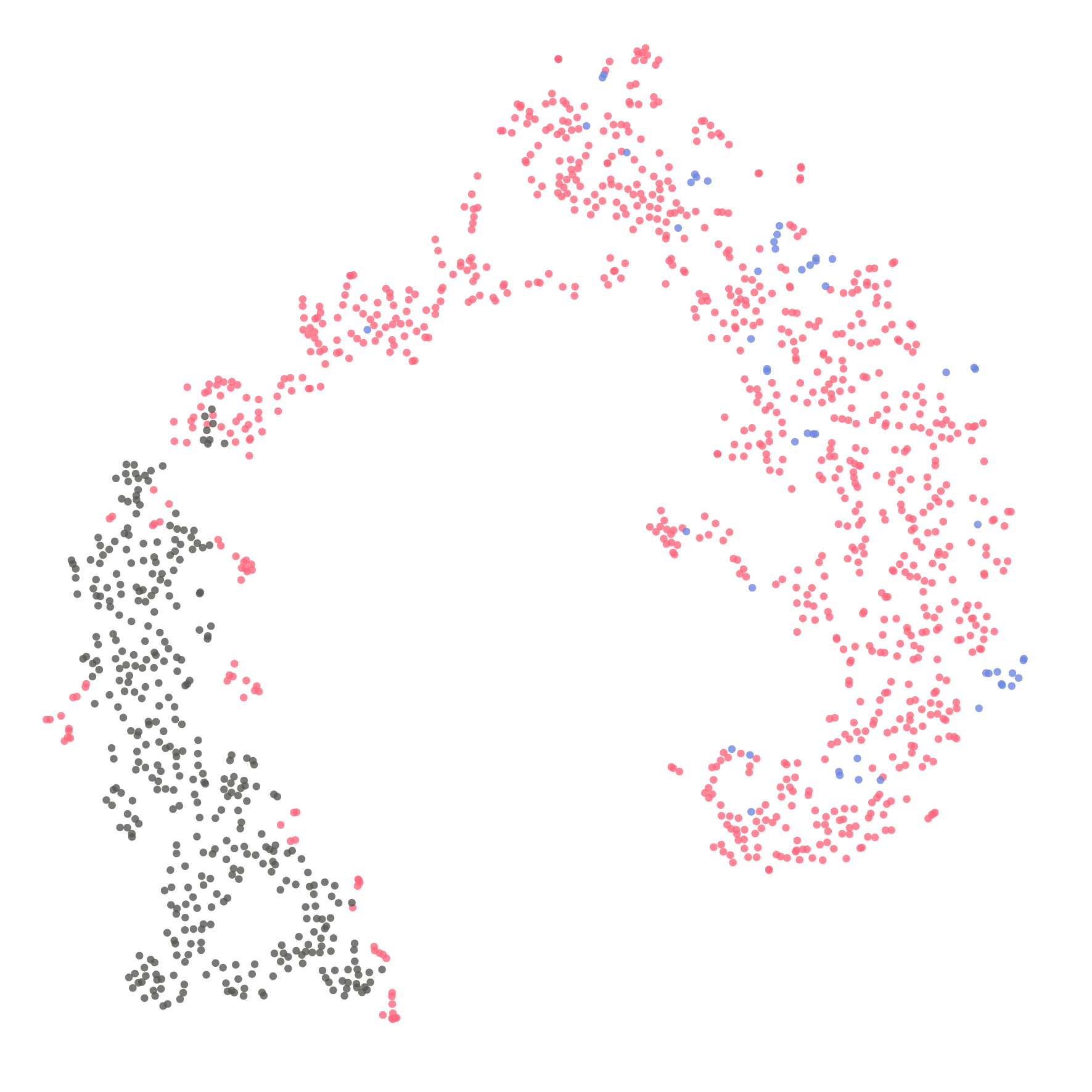}
	\caption{Spectral (visible to NIR) features \label{fig:vis-nir}}
	\end{subfigure}
	\caption{t-SNE projections of hand-crafted representations of $64\times64$ pixel hyperspectral patches from the Pavia University, Houston University and Toulouse data sets. (a) corresponds to the 2D projection of the spatial features only, \textit{i.e.} only the Gabor filters were used to represent the data and then the t-SNE projection was performed, and (b) corresponds to the 2D projection of the spectral features only, \textit{i.e.} only the spectral indices were used to represent the data and then the t-SNE projection was performed. Moreover, only the smallest spectral domain, \textit{i.e.} the spectral domain of the Pavia University image which covers the 0.4 µm - 0.86 µm range, was used. \label{fig:proj2}}
\end{figure}

\clearpage

\begin{figure}[h]
	\center
	\includegraphics[width=0.8\textwidth]{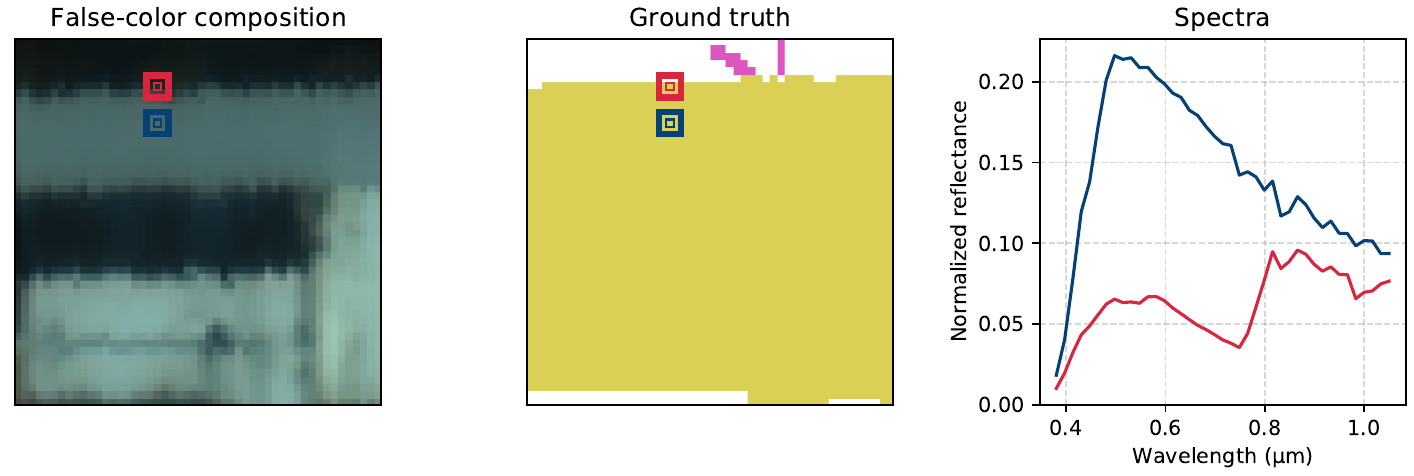}
		
	\includegraphics[width=0.8\textwidth]{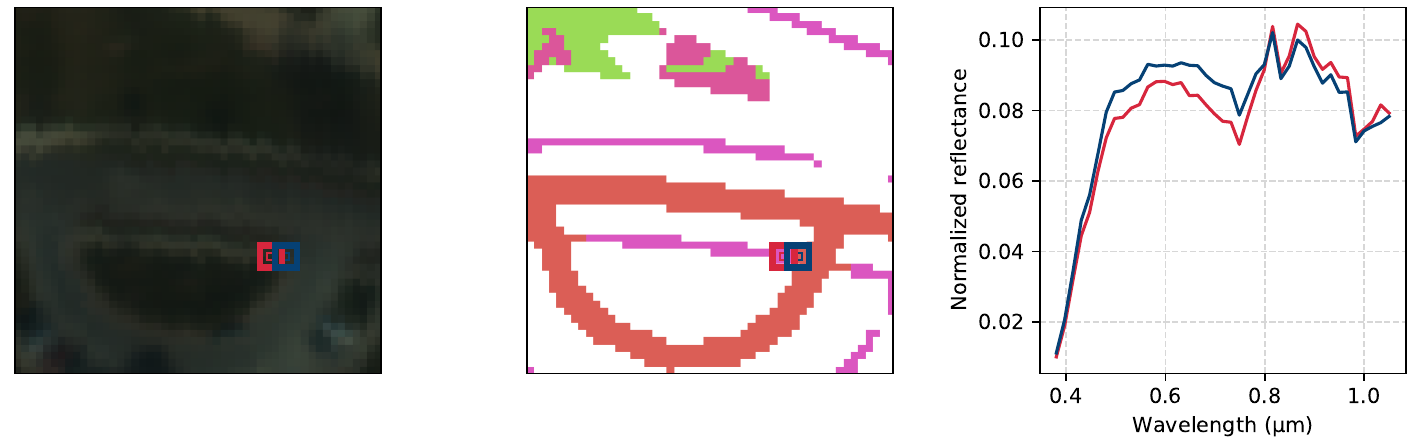}
		
	\includegraphics[width=0.8\textwidth]{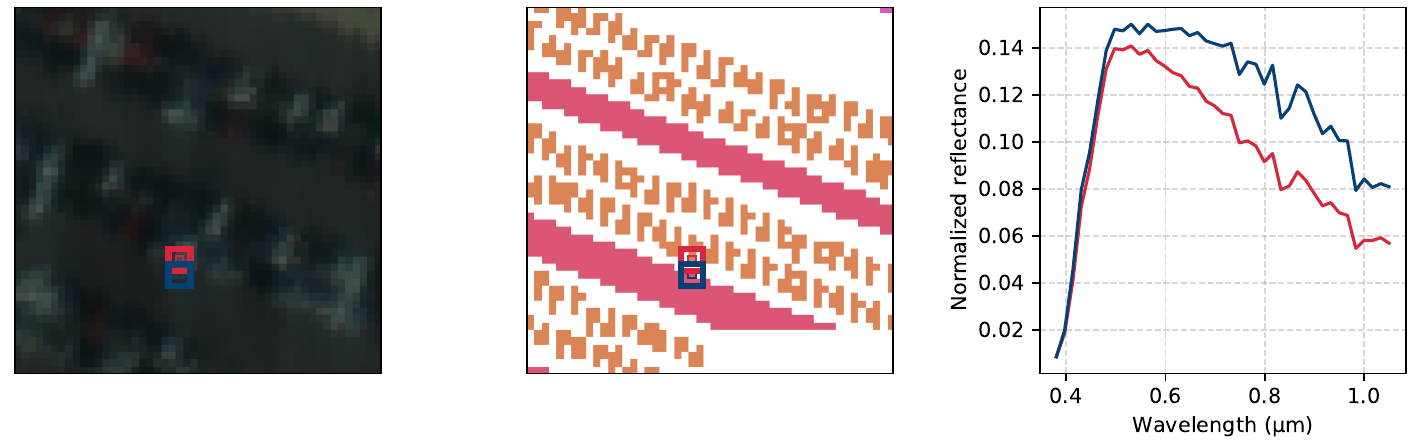}
	
	\caption{Illustration of noisy labels in the Houston university data set. The first row shows an example of two spectra labeled as \textit{Non-residential buildings}. The pixel at the edge of the roof is mixed with the ground vegetation, which provides erroneous spectral information. The second and third rows show neighboring pixels labeled with different classes though they are actually a mixed of both classes. \label{fig:noise}}

\end{figure}

\begin{figure}[h]
	\center
	\includegraphics[width=0.2\textwidth]{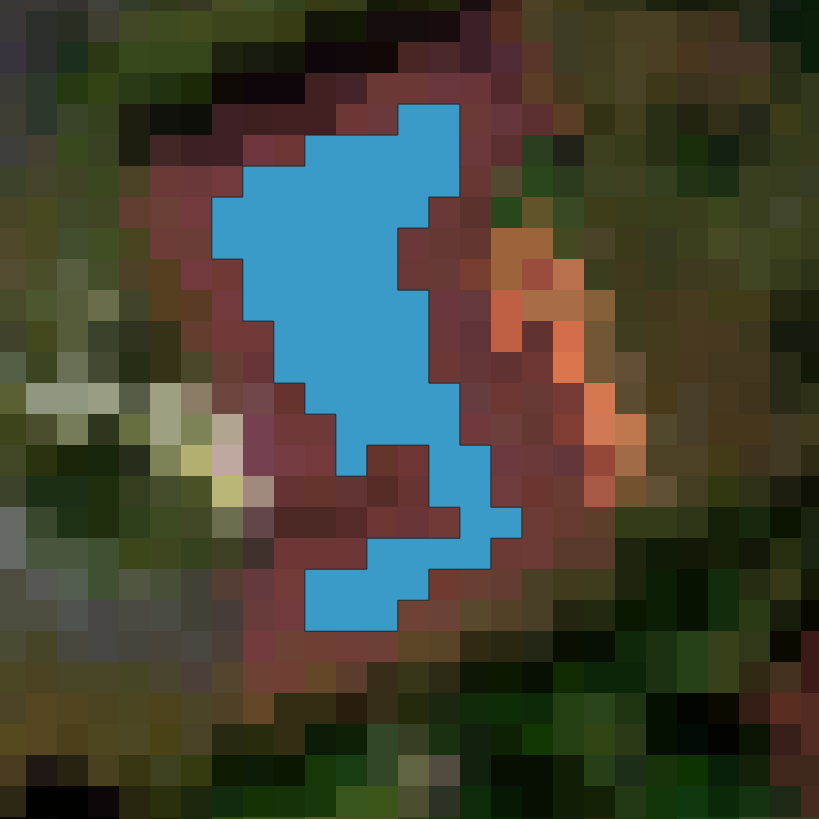}
	\hspace{0.5cm}
	\includegraphics[width=0.2\textwidth]{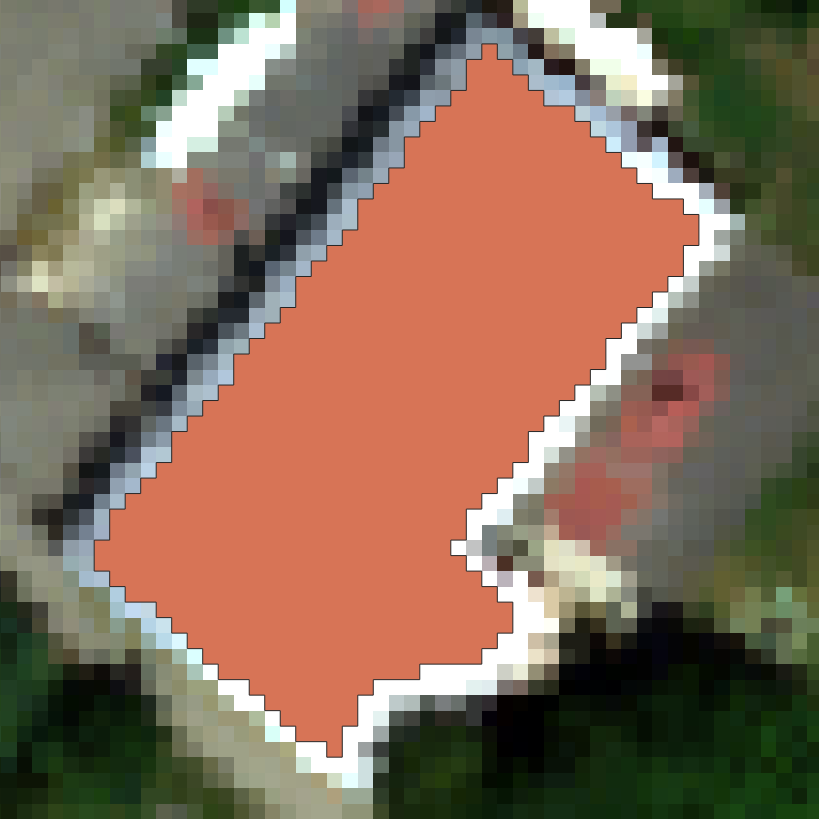}
	\hspace{0.5cm}	
	\includegraphics[width=0.2\textwidth]{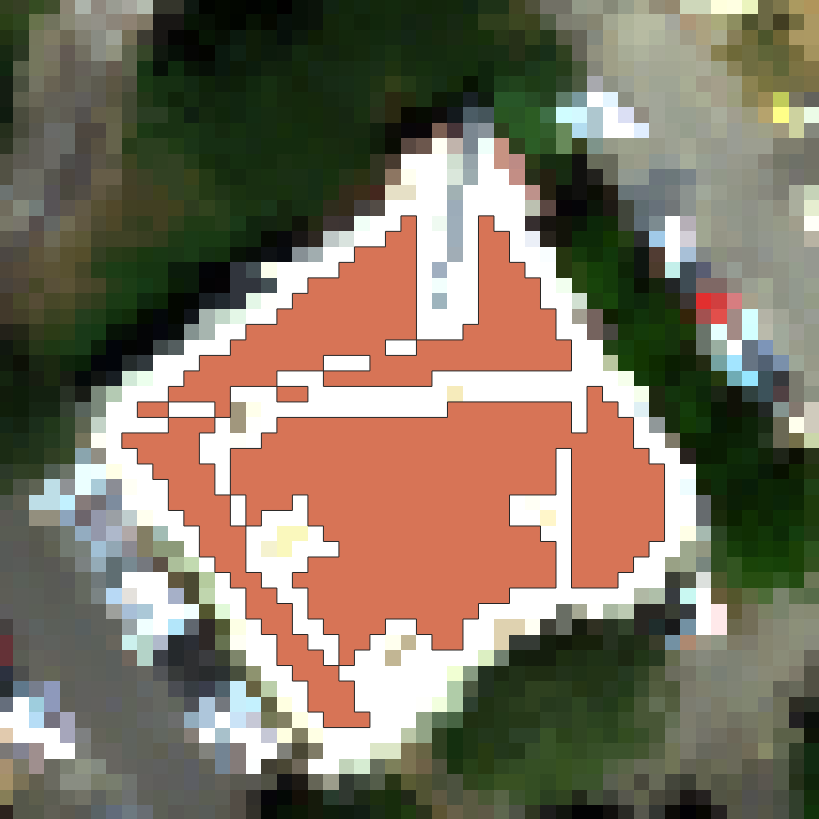}
	
	\caption{Examples of annotations in the Toulouse data set. We paid attention to ignore mixed pixels such as pixels at edges of roofs or pixels mixed with small objects such as pipes on roofs. \label{fig:examples_tlse}}
\end{figure}

\end{appendices}

\end{document}